\pdfoutput=1

\PassOptionsToPackage{table}{xcolor}

\documentclass{article}

\usepackage{times}
\usepackage{latexsym}

\usepackage[T1]{fontenc}

\usepackage[utf8]{inputenc}

\usepackage{microtype}

\usepackage{inconsolata}

\usepackage{graphicx}

\usepackage{subfigure}
\usepackage{booktabs} 

\usepackage{hyperref}

\usepackage[preprint]{acl}

\usepackage{amsmath}
\usepackage{amssymb}
\usepackage{mathtools}
\usepackage{amsthm}

\usepackage[table]{xcolor}

\usepackage[capitalize,noabbrev]{cleveref}

\theoremstyle{plain}

\theoremstyle{definition}

\theoremstyle{remark}

\usepackage[textsize=tiny]{todonotes}

\usetikzlibrary {shapes.geometric}

\def\shortpar#1{\textbf{#1.}}

\long\def\devour#1{}

\newcounter{notecounter}
\newcommand{\enotesoff}{\long\gdef\enote##1##2{}}
\newcommand{\enoteson}{\long\gdef\enote##1##2{{
\stepcounter{notecounter}
{\large\bf
\hspace{0cm}\arabic{notecounter} $<<<$ ##1: ##2
$>>>$\hspace{1cm}}}}}

\enoteson
\enotesoff

\title{Understanding Gated Neurons in Transformers from Their Input-Output Functionality}

\author{Sebastian Gerstner \and Hinrich Schütze \\
  Center for Information and Language Processing (CIS), LMU Munich, \\
  and Munich Center for Machine Learning (MCML),\\
  Germany \\
  \texttt{sgerstner@cis.lmu.de}
  }

\begin{document}

\maketitle

\begin{abstract}
Interpretability researchers have attempted to understand
MLP neurons of language models based on both the contexts in
which they activate and their output weight vectors.  They
have paid little attention to a complementary aspect: the
interactions between input and output.  For example, when
neurons detect a direction in the input, they might add much
the same direction to the residual stream (``enrichment
neurons'') or reduce its presence (``depletion neurons'').
We address this aspect by examining the cosine similarity
between input and output weights of a neuron.  We apply our
method to 12 models and find that enrichment neurons
dominate in early-middle layers whereas later layers tend
more towards depletion.  To explain this finding, we argue
that enrichment neurons are largely responsible for
enriching concept representations, one of the first steps of
factual recall.  Our input-output perspective
is a complement to activation-dependent analyses
and to approaches that treat input and output separately.
\end{abstract}



\enote{hs}{the abstract is very thin on tangible
results. one strong tangible result you could add: talk
about the commonality you found wrt to enrichtment/depletion
across all models you analyzed}

\enote{hs}{``input manipulation'' could still
be more integrated into the text}

\section{Introduction}

Despite recent progress in interpretability,
there is still much that is unclear
about how transformer-based 
\cite{2017_Vaswani} large language models (LLMs) achieve
their impressive performance.
Prior work has
addressed the interpretation of MLP sublayers,
and we follow this line of research.
Some of this work analyzes neurons based
only on the contexts in which they activate
\cite{voita-etal-2024-neurons}
or based only on their output
weights\footnote{We use ``weight'' to refer to a
weight vector, not a scalar.}
\cite{2024_Gurnee}.
In contrast, we put the input-output (IO)
functionality of neurons in the center of our analysis,
and classify neurons according to the interactions between
input and output weights.
We focus on gated activation functions \cite{2020_Shazeer},
which are used in recent LLMs like OLMo, Llama and Gemma.

\begin{figure}
	\centering
	\includegraphics
	[width=\linewidth]
	{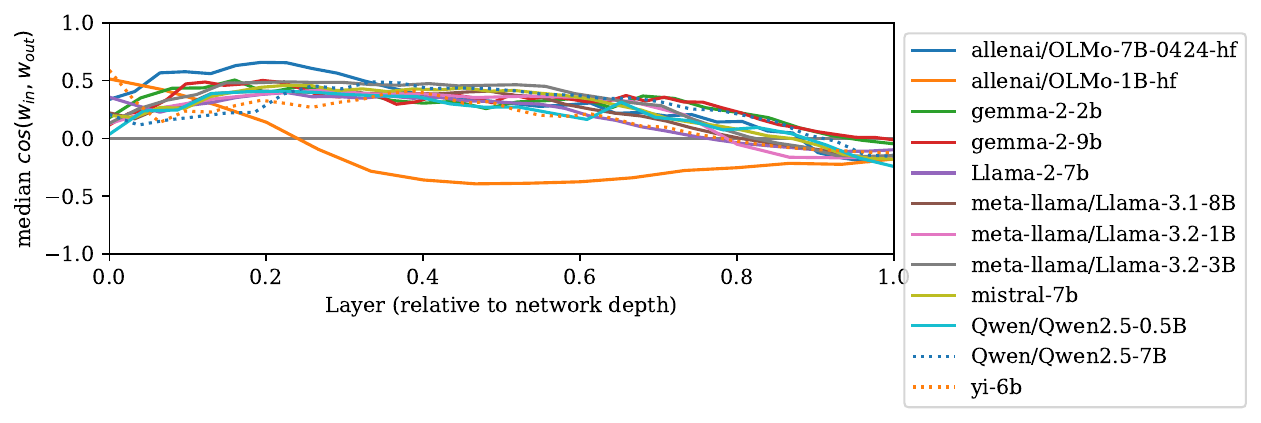}
	\caption{
	Median of $\cos(w_{\text{in}},w_{\text{out}})$ by layer (x-axis)
	for 12 models.
	For all models, the value is positive in
the beginning
and negative in the end,
        indicating that early-middle layers ``enrich'' the
	residual stream whereas
later layers tend
more towards depletion.}
	\label{fig:medians}
\end{figure}



\shortpar{Theoretical framework}
Following \citet{2021_Elhage},
our view of the Transformer architecture is centered on the residual (a.k.a. skip) connections between sublayers:
they form the \textit{residual stream},
and the individual units (such as MLP neurons) progressively update it,
until it is multiplied by the unembedding matrix $W_U$ to produce next-token logits.
The information contained in the residual stream is represented as a high-dimensional vector (of dimension $d_{\text{model}}$).
Individual model units \textit{read} from the residual stream
and then update it by \textit{writing} (adding) other vectors to it.
In the case of an MLP neuron, it 
detects certain directions in the residual stream
(i.e., whether the current residual stream vector at least approximately points in one of these directions in model space),
corresponding to its weight vectors on the input side;
and then writes to a certain direction, corresponding to its output weight vector.

A semantic intepretation is that a neuron
detects a \textit{concept} in the residual stream
(for example, an intermediate guess about the next token),
and in turn also writes a concept.
This semantic interpretation is not a necessary assumption for our neuron classification,
but is helpful for building intuition and interpreting results.


\shortpar{Theoretical contribution}
These theoretical reflections naturally lead to our research question:
\textbf{What is the relationship between what a neuron reads and what it writes?}
We address this question by computing the
\textbf{cosine similarity of input and output weights},
focusing on
gated activation functions.

Specifically,
with gated activation functions,
each neuron has three weight vectors: the linear input, gate, and output weight vectors.
When the output weight is similar enough to (one of) the detected directions,
we speak of \textbf{input manipulation},
as opposed to \textbf{orthogonal output} neurons
which write to
directions not detected in the input.
Intuitively, input manipulator neurons \textit{manipulate} the concept that they detect.
As special cases of input manipulation,
we define \textbf{enrichment}
and \textbf{depletion} neurons -- neurons that
detect a direction 
and then
add it
to / remove it from the residual stream.
We
present a complete taxonomy of neuron IO functionalities in
\cref{sec:ss:prototypes}.
See \cref{fig:prototypes} for a visualization.

\shortpar{Empirical study}
We apply our method to 12 LLMs.
We find that,
for all of these models,
a large proportion of neurons are input manipulators.
In particular,
we find that
enrichment neurons dominate in early-middle layers of all
models
whereas
later layers tend
more towards depletion. See
\cref{fig:medians}.


We also present examples for the six major IO functionalities.
We find that many neurons
have the property of \textbf{double checking}:
The
two reading weight vectors ($w_{\text{gate}}$ and $w_{\text{in}}$) are
approximately orthogonal, but still intuitively represent
the same concept.

\shortpar{Explaining the results}
Our finding of different IO functionalities in different layers
echoes the ``stages of inference'' framework \cite{lad2024remarkablerobustnessllmsstages}.
We hypothesize a correspondence:
enrichment neurons may be responsible for ``feature engineering''
and depletion neurons for ``residual sharpening''.

We also provide a theoretical account of the double checking phenomenon.
The usefulness of double checking explains the fact that
many neurons have approximately orthogonal gate and input weights.





\shortpar{Contributions}
(i)
We develop a parameter-based (and therefore efficient)
method to
investigate neuron IO functionalities for gated activation functions
(\cref{sec:theory}).
(ii)
Across 12 models,
we find that
enrichment neurons dominate in early-middle layers of all
models
whereas
later layers tend
more towards depletion
(\cref{fig:medians}).
(iii)
We define two novel concepts helpful in
understanding neuron functionality:
\textit{input manipulation}
and
\textit{double checking}.
(iv)
We find that many neurons are input manipulators
(\cref{sec:exp}),
which makes our classification scheme useful for understanding them.
(v)
We present examples for
the six major IO functionalities,
showing how the IO perspective complements other neuron analysis methods
(\cref{sec:cs}).
(vi)
We propose theoretical explanations for some of these results
(\cref{sec:discussion}).




\section{Related Work}\label{sec:background}\label{sec:ss:related}

There is a large body of work on interpretability of
transformer-based LLMs.
\citet{2021_Elhage} introduce the notion of residual stream.
\citet{2020_LogitLens}, \citet{2023_Belrose} propose to interpret residual stream states as intermediate guesses about the next token;
\citet{pmlr-v235-rushing24a} discuss this as the \textit{iterative inference hypothesis}.
On a similar note, many works hypothesize that directions in model space can correspond to concepts;
\citet{pmlr-v235-park24c} discuss this as the \textit{linear representation hypothesis}.
\citet{lad2024remarkablerobustnessllmsstages} define \textit{stages of inference}.
\citet{geva-etal-2023-dissecting} explain how LLMs recall facts;
a crucial early step is \textit{representation enrichment},
which may be
related to our \textit{enrichment neurons} (see \cref{sec:discussion-enrichment}).
Similar to our work,
\citet{2024_Elhelo}
investigate input-output functionality of heads (instead of neurons).


Much research has attempted to understand individual
neurons.
\citet{geva-etal-2021-transformer}
present them as a key-value memory.
Other neuron analysis work includes \citep{2023_Miller,2024_Niu}.
The focus on individual neurons has been criticized.
\citet{2018_Morcos} find
that in good models, neurons are not monosemantic
(but for image models, not LLMs).
\citet{2022_Millidge} compute a singular value decomposition (SVD) of layer weights and often find interpretable directions that do not correspond to individual neurons.
\citet{2022_Elhage} argue that interpretable features 
are non-orthogonal directions in model space
and can be superposed.
This corresponds to sparse linear combinations of neurons in MLP space.
Taking the middle ground,
\citet{2023_Gurnee} 
argue that interpretable features correspond to sparse combinations of neurons,
but this includes 1-sparse combinations, i.e., individual neurons.

Several works classify neurons based on the
\textbf{contexts} in which they activate \citep{voita-etal-2024-neurons,2024_Gurnee}.
For example, \citet{voita-etal-2024-neurons} find \textit{token detectors} that
suppress
repetitions. \citet{2024_Gurnee} also
define \emph{functional roles} of neurons based on
their \textbf{output} weight vector, such
as \textit{suppression neurons} that suppress a specific set
of tokens.  They note that suppression neurons seem to
activate ``when it is plausible but not certain that the
next token is from the relevant set''.
\citet{2024_Stolfo} also investigate some output-based neuron classes.

Researchers have paid less attention to the input-output
perspective. \citet{2024_Gurnee} compute cosine
similarities between input and output weights for GPT-2
\citep{radford2019language}, but do not interpret their results.
\citet{2022_Elhage} 
mention the idea of input-output analysis
(negative cosines between input and output weights ``may also
be mechanisms for conditionally deleting information'',
footnote 7), but do not follow up on this remark.  Note also
that input-output analysis for gated activation functions
adds complexity because, in addition to input and output
weight vectors, the gating mechanism is crucial for IO functionality.

\section{Gated activation functions}\label{sec:ss:swiglu}

In our neuron classification we assume
\textit{gated activation functions} like SwiGLU or GeGLU \cite{2020_Shazeer}.
In this section, we describe
definition (\cref{sec:swiglu-def}) and properties (\cref{sec:swiglu-properties})
of these functions.
Gated activation functions are used widely, e.g., 
OLMo \cite{groeneveld-etal-2024-olmo}
and Llama
\cite{llama}
use SwiGLU,
and Gemma
\cite{gemma_2024}
uses GeGLU.

The following description focuses on SwiGLU.
GeGLU replaces Swish with GeLU, but is otherwise identical.
For a visualization of a SwiGLU neuron, see \cref{fig:swiglu} in \cref{ap:swiglu}.

\subsection{Definitions}\label{sec:swiglu-def}
To keep our description simple,
we ignore bias terms and layer norm parameters.
(Some models, like OLMo, lack these anyway.)
We describe single neurons as opposed to whole MLP layers.

\def\xln{x_{\text{norm}}}
We denote by $x_{\text{mid}}$ the state of the residual stream before the MLP, and
by $\xln := \text{LN}(x_{\text{mid}})$ its layer normalization.
We say that a \textbf{direction} $v\in\mathbb{R}^d$
is \textbf{present} (positively) in a vector
$x\in\mathbb{R}^d$ if $x\cdot v \gg 0$.

Traditional activation functions like ReLU take a single scalar as argument: $\mbox{ReLU}(x_{\text{in}})$.
In contrast, a \emph{gated activation function}
like SwiGLU
takes two arguments:
\[\mbox{SwiGLU}(x_{\text{gate}}, x_{\text{in}}) = \mbox{Swish}(x_{\text{gate}}) \cdot x_{\text{in}}.\]

To compute the scalars $x_{\text{gate}}$ and $x_{\text{in}}$,
each neuron has a \textbf{linear input} weight vector
$w_{\text{in}}$
\enote{hs}{no space? footnote{
also known as ``up-projection''
}}
and a \textbf{\text{gate}} weight vector $w_{\text{gate}}$ of dimension $d_{\text{model}}$.
We
refer to these two weight vectors as
the \textbf{reading weights}.
Then $x_{\text{gate}}$ is defined as $w_{\text{gate}} \cdot \xln$,
and $x_{\text{in}}$ as $w_{\text{in}} \cdot \xln$.



Finally,
the product of
$\mbox{SwiGLU}(x_{\text{gate}}, x_{\text{in}})$
and  the \textbf{output} weight
vector,
$w_{\text{out}}$
\enote{hs}{no space?
footnote{
also known as ``down-projection''
}},
is 
added to the residual
stream.



\subsection{Properties}\label{sec:swiglu-properties}
There are three properties of
gated activation functions that are key for
understanding IO functionality.

\textbf{Positive vs negative activation.}
Strong activations can be either positive or negative.
If $w_{\text{gate}}\cdot\xln \gg 0$ and
$w_{\text{in}}\cdot\xln \gg 0$,
the activation is
strongly positive.
If $w_{\text{gate}}\cdot\xln \gg 0$ and
$w_{\text{in}}\cdot\xln \ll 0$,
the activation is
strongly negative.
So,
depending on the context,
a given gated activation neuron can either add
the output weight vector to the residual stream or subtract it.

\textbf{Negative values of Swish.}
Swish and GeLU are often seen as essentially ReLU.
However, we found clearly different cases (see \cref{sec:cs}).
$w_{\text{gate}}\cdot x_{\text{norm}}$ can be \textbf{weakly negative},
i.e., negative but close to zero.
In this case its image under Swish is also weakly negative.
This leads to a negative activation if $w_{\text{in}}$ is present positively and positive otherwise.

\label{sec:ss:symmetry}
\textbf{Symmetry.}
Switching the signs of both $w_{\text{in}}$ and $w_{\text{out}}$ preserves
IO behavior.

\section{Method}
\label{sec:ss:prototypes}\label{sec:theory}

\begin{figure}\centering
	\includegraphics
	[width=\linewidth]
	{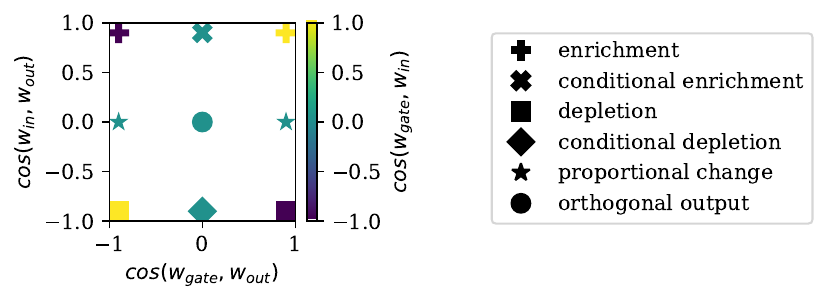}
	\caption{We define six input-output
	functionality classes  or \textbf{IO classes} of
        gated activation neurons based on  collinearity
	and orthogonality of their
        linear input, gate and output
	weight vectors. For example, depletion neurons remove the
	direction of the gate vector from the residual
	stream. Examples shown are prototypical.}
	\label{fig:prototypes}
\end{figure}

\begin{table*}\centering
    \begin{tabular}{c|cccc}
        & \multicolumn{2}{c}{$|\cos(w_{\text{gate}},w_{\text{out}})| \gg 0$}  &\multicolumn{2}{c}{$\cos(w_{\text{gate}},w_{\text{out}}) \approx 0$}\\
            $\cos(w_{\text{in}},w_{\text{out}})$ \\     

        \hline
        
\rowcolor{gray!25}       $ \gg 0$
        & \multicolumn{2}{c}{\textbf{enrichment}} 
        
        & \multicolumn{2}{c}{\textbf{conditional
        enrichment}}\\
	&  {\scriptsize $|\cos(w_{\text{gate}},w_{\text{in}})| \gg 0$} &
        {\scriptsize $|\cos(w_{\text{gate}},w_{\text{in}})| \approx 0$}&
          {\scriptsize $|\cos(w_{\text{gate}},w_{\text{in}})| \approx 0$}
        &  {\scriptsize $|\cos(w_{\text{gate}},w_{\text{in}})| \gg 0$}\\
        &typical & atypical
        &typical & atypical\\
\rowcolor{gray!25}          $ \ll 0$
      & \multicolumn{2}{c}{\textbf{depletion}} 
        & \multicolumn{2}{c}{\textbf{conditional depletion}}\\
	& {\scriptsize $|\cos(w_{\text{gate}},w_{\text{in}})| \gg 0$} &
        {\scriptsize $|\cos(w_{\text{gate}},w_{\text{in}})| \approx 0$}&
          {\scriptsize $|\cos(w_{\text{gate}},w_{\text{in}})| \approx 0$}
        &  {\scriptsize $|\cos(w_{\text{gate}},w_{\text{in}})| \gg 0$}\\
        &typical & atypical
        &typical & atypical\\
\rowcolor{gray!25}          $ \approx 0$
      & \multicolumn{2}{c}{\textbf{proportional change}} 
        & \multicolumn{2}{c}{\textbf{orthogonal output}}\\
        & {\scriptsize $|\cos(w_{\text{gate}},w_{\text{in}})| \approx 0$}
        &  {\scriptsize $|\cos(w_{\text{gate}},w_{\text{in}})| \gg 0$}&&\\
        &typical & atypical
\\
    \end{tabular}
	\caption{Our six IO classes, in \textbf{boldface}. Five of them have
        ``atypical'' variants.
    We use a threshold of 0.5 (resp.\ -0.5) to distinguish
$\cos()\approx 0$ from $|\cos()|\gg 0$.}
    \label{tab:alldefs}
\end{table*}

We now describe 
how we investigate input-output functionalities
of gated neurons,
based on their weights only.

\enote{hs}{cut for space
(This is inspired by one of the neurons we found in \cref{sec:cs}.)
}

\subsection{Intuition}
As a running example, we consider what a neuron would do to a residual stream state representing the next-token prediction \textit{review}.
\enote{hs}{
cut for space (with a running example, there is no claim
that it is a general represenetation of all cases).
Note however that the residual stream can and usually does
represent more abstract (perhaps non-interpretable) concepts
and/or multiple concepts at a time.
}

Before we introduce our method,
let us consider a simpler case to develop our intuition:
non-gated activation functions like ReLU (see also
\citet{2024_Gurnee}). 
Here, a neuron detects just one direction, determined by its input weight $w_{\text{in}}$ (say \textit{review}).
(Given
$x_{\text{norm}}$, the activation depends only on $x_{\text{norm}} \cdot w_{\text{in}}$,
and is positive whenever this is positive.)
Roughly, we can distinguish three  cases:
the neuron output (determined by $w_{\text{out}}$) can be
similar to the input direction (in our case, \textit{review}: we call this \textit{enrichment}),
different (we call this \textit{orthogonal output}),
or roughly opposite (in our case, ``minus \textit{review}'': we call this \textit{depletion}).
In terms of weights, these cases correspond to $cos(w_{\text{in}}, w_{\text{out}})$ being close to 1, close to 0, or close to -1.

Note that a neuron could also detect ``minus \textit{review}''
(i.e., ``\textit{review} is not the next token''),
and enrich or deplete that direction.

\subsection{Extension to gated activation functions}
In this paper, we consider gated activation functions.
Here, a neuron detects two directions ($w_{\text{gate}}$ and
$w_{\text{in}}$), not one;
so there are more cases to consider.
Luckily, the symmetry property (see \cref{sec:ss:symmetry}) simplifies the analysis:
a neuron's behavior does not change if we switch the signs of
both $w_{\text{in}}$ and $w_{\text{out}}$.
This implies that \textit{the sign of $\cos(w_{\text{gate}},w_{\text{out}})$ does not matter}.

Accordingly, we define six IO classes, depending on
$cos(w_{\text{in}},w_{\text{out}})$ (three rows: positive, negative, or zero)
and $|cos(w_{\text{gate}},w_{\text{out}})|$ (two columns: positive or zero).
Although there is a third cosine similarity --
$cos(w_{\text{gate}},w_{\text{in}})$ --
 this similarity is determined by the two others in
 prototypical cases.
We will consider these prototypical cases first.

\subsection{Prototypical cases}
See \cref{tab:alldefs} for an overview of all cases and
\cref{fig:prototypes} for a visualization.
We also encourage the use of the interactive visualization in supplementary.
\enote{sg}{Don't forget to actually include it it the supplementary material!}

For the prototypical cases we assume the cosines are
$\approx$ 1,
$\approx$ $-1$
or $\approx$ 0.
In these cases, knowing two of the cosine similarities implies knowing the third one:
If
$w_{\text{gate}}$ and $w_{\text{in}}$
are collinear,
then
 $w_{\text{out}}$ has the
same cosine similarity with both (up to sign). Conversely,
if $w_{\text{gate}}$ and $w_{\text{in}}$ are orthogonal, $w_{\text{out}}$
cannot be collinear to both, and in fact,
$\cos(w_{\text{gate}},w_{\text{out}})^2+\cos(w_{\text{in}},w_{\text{out}})^2 \leq 1$,
with equality when $w_{\text{out}}$ is in the space spanned by
$w_{\text{gate}}$ and $w_{\text{in}}$.

We first focus on 
on textbf{enrichment and depletion}:
$\cos(w_{\text{in}},w_{\text{out}}) \approx \pm 1$.
The gate vector can be collinear as well, i.e.,
$\cos(w_{\text{gate}},w_{\text{out}})\approx\pm1$
(leftmost ``typical'' column).
In this case, all three vectors are
approximately in a one-dimensional subspace,
so the neuron detects one direction and writes to the same direction, up to sign.
The sign is relevant:
Assume $x_{\text{norm}}$ represents the token \textit{review} and $w_{\text{gate}}$ detects that direction, so that the neuron activates.
If $cos(w_{\text{in}},w_{\text{out}})\approx 1$
($w_{\text{in}},w_{\text{out}}$ also lie in the \textit{review} subspace, and both have the same orientation),
the neuron will again write \textit{review}.
We call this (typical) \textbf{enrichment}.
On the other hand, if $cos(w_{\text{in}},w_{\text{out}})\approx -1$
(they again lie in the \textit{review} subspace but have different orientations),
the neuron will write
``minus \textit{review}''.
We call this (typical) \textbf{depletion}.\footnote{
We prefer these terms to alternatives like \textit{increase / reduction}
because in practice output directions will not be exactly the same as the reading directions.
See \cref{sec:discussion-enrichment}.
}
The same neurons can also get a \textit{weak negative} activation if $-w_{\text{gate}}$ (``minus \textit{review}'') is weakly present in the residual stream.
In this case, Swish
has a negative value (\cref{sec:swiglu-properties})
and the enrichment neuron  writes ``plus \textit{review}''
to the residual stream
and the depletion neuron ``minus \textit{review}''.


Next we consider
\textbf{conditional enrichment} and \textbf{conditional depletion}:
$w_{\text{in}}$ and $w_{\text{out}}$ are roughly collinear and
$w_{\text{gate}}$ is orthogonal to them.
Consider the example that $w_{\text{in}}, w_{\text{out}}$ correspond to the \textit{review}
direction and $w_{\text{gate}}$ to 
``verb expected as next token''.
The neuron will only activate \emph{conditional on
$w_{\text{gate}}$ being present in the residual stream}
(here: verb expected).
If $\pm w_{\text{in}}$ (``plus'' or ``minus'' \textit{review}) is \textit{also}
present in the residual stream,
then $\pm w_{\text{out}}$ (``plus'' or ``minus'' \textit{review})
will be added to the residual stream.
For this scenario, we define a 
(typical) \textbf{conditional enrichment}
neuron as one
with $\cos(w_{\text{in}},w_{\text{out}})\approx 1$;
this neuron will enrich the residual stream with $w_{\text{in}}$ if $w_{\text{in}}$ is present
and with $-w_{\text{in}}$ if $-w_{\text{in}}$ is present
(``plus'' \textit{review} leads to more of ``plus'' \textit{review},
and ``minus'' \textit{review} leads to more of ``minus'' \textit{review}).
Conversely, we define a
(typical) \textbf{conditional depletion} neuron as one that
depletes $\pm w_{\text{in}}$ (whichever was present) from the residual stream:
``plus'' \textit{review} leads to ``minus'' \textit{review} and vice versa.
As before, if $-w_{\text{gate}}$ is weakly present in the residual stream
(there is a weak expectation that the next token is not a verb),
Swish yields a negative value;
so in this situation conditional enrichment and depletion
neurons switch their behaviors; e.g.,
for a conditional enrichment neuron ``plus'' \textit{review} will lead to ``minus'' \textit{review}.


Turning to the bottom part of
\cref{tab:alldefs},
we define a
(typical) \textbf{proportional change neuron}
as one whose $w_{\text{out}}$ is in the same
one-dimensional subspace as $w_{\text{gate}}$,
but is orthogonal to $w_{\text{in}}$.
(This implies that $w_{\text{gate}}$ and $w_{\text{in}}$ are orthogonal.)
Take the case where
$w_{\text{gate}},w_{\text{out}})$ are 
represent \textit{review}  and
$w_{\text{in}}$  ``verb expected''.
If $w_{\text{gate}}$ (\textit{review}) is present in the residual stream,
then the neuron writes a \emph{positive or negative} multiple of \textit{review} to the residual stream.
This multiple is proportional to the presence of $w_{\text{in}}$ (``verb expected'') in the residual stream:
If a verb is expected, the neuron writes \textit{review}, if not, it writes ``minus \textit{review}''.


All of the above neuron types are \textbf{input manipulators}:
they write to one of the directions they detect.
Our final category is the negation of this:
We define an
\textbf{orthogonal output
neuron} as one
whose output weight vector is
orthogonal to both reading weight vectors.
If $w_{\text{gate}}$ and $w_{\text{in}}$ are also orthogonal to each other,
then such a neuron 
defines an interaction of three completely different meaning
components.

\subsection{General case: Typical vs atypical functions}
Many cosines will not be close to 0 or $\pm 1$.
For example,
such a neuron may write a concept different from but
semantically related to the one it detects
(say, \textit{Ireland} -> \textit{Dublin}) and thus be
be similar to an enrichment neuron
in terms of weight vector geometry.

For this general case, this paper explores three options to understand
neuron IO functionalities
at different levels of granularity:
(1)
Classify neurons according to the closest prototypical case.
(2)
Plot the marginal distributions of the three cosine similarities.
(3)
Place neurons in a plot analogous to \cref{fig:prototypes}, based on their three weight cosines.

For (1),
 we need two refinements. (i) We need a threshold $\tau$
for 
counting a cosine similarity as clearly different from
zero.
In this paper, we set $\tau = 0.5$,
a relatively permissive cutoff
that we believe
gives rise to a more informative classification of neurons.

\enote{hs}{reinstate if there is space
Indeed, this threshold of
$0.5$ is a lower bar than the $\frac{\sqrt{2}}{2}\approx
0.7$ corresponding to an angle of $\pi/4 =
45^\circ$.
\citet{2024_Gurnee} 
find that few cosine
similarities are close to $\pm 1$.
Our results
in \cref{sec:ss:res-wcos} confirm this.}

(ii) 
$\cos(w_{\text{in}},w_{\text{gate}})$ may not always ``match'' the other two cosine similarities;
e.g., the two reading weights
may be orthogonal, but
$w_{\text{out}}=\frac{1}{\sqrt{2}} w_{\text{gate}} + \frac{1}{\sqrt{2}} w_{\text{in}}$;
then both cosine similarities are
$\frac{1}{\sqrt{2}} > 0.5$.
We are mainly interested in IO behavior
rather than comparing the two reading weights,
so we classify such cases based
on $\cos(w_{\text{in}},w_{\text{out}})$ and
$\cos(w_{\text{gate}},w_{\text{out}})$.
To signal the ``mismatch'' of $\cos(w_{\text{in}},w_{\text{gate}})$,
we prepend \textbf{atypical}
to the category's name.
In the above example, we will speak of
an atypical enrichment neuron.
In 
\cref{fig:prototypes}, the atypical classes
share their position with typical classes, but differ in color.

\cref{tab:alldefs} shows all atypical (and typical) classes.

\section{IO functionalities by layers}\label{sec:exp}
\label{sec:ss:res-wcos}

\enote{sg}{TODO here or above:
random baseline}


We conduct our study on
12 models:
Gemma-2-2B, Gemma-2-9B \cite{gemma_2024},
Llama-2-7B, Llama-3.1-8B, Llama-3.2-1B, Llama-3.2-3B \cite{llama},
OLMo-1B, OLMo-7B-0424 \cite{groeneveld-etal-2024-olmo}, 
Mistral-7B \cite{jiang2023mistral7b},
Qwen2.5-0.5B, Qwen2.5-7B \cite{qwen2},
Yi-6B \cite{ai2025yiopenfoundationmodels}.
These models use SwiGLU,
except for Gemma, which uses GeGLU.
For each model,
we classify the MLP neurons based on the cosine similarities of the three weight vectors,
as described in \cref{sec:theory}.

Here we describe the results for \textbf{Llama-3.2-3B}.
They are representative of the general trends we observe.
\cref{ap:hr} in the appendix contains the plots for all models.

We progress from (i) the coarse-grained version of our
method, with discrete classes, to (ii)
the marginal distributions of each cosine similarity, to (iii)
fine-grained scatter plots showing all individual neurons.

\subsection{Discrete classes}\label{sec:discrete}
\begin{figure}
	\centering
	\includegraphics
	[width=\linewidth]
	{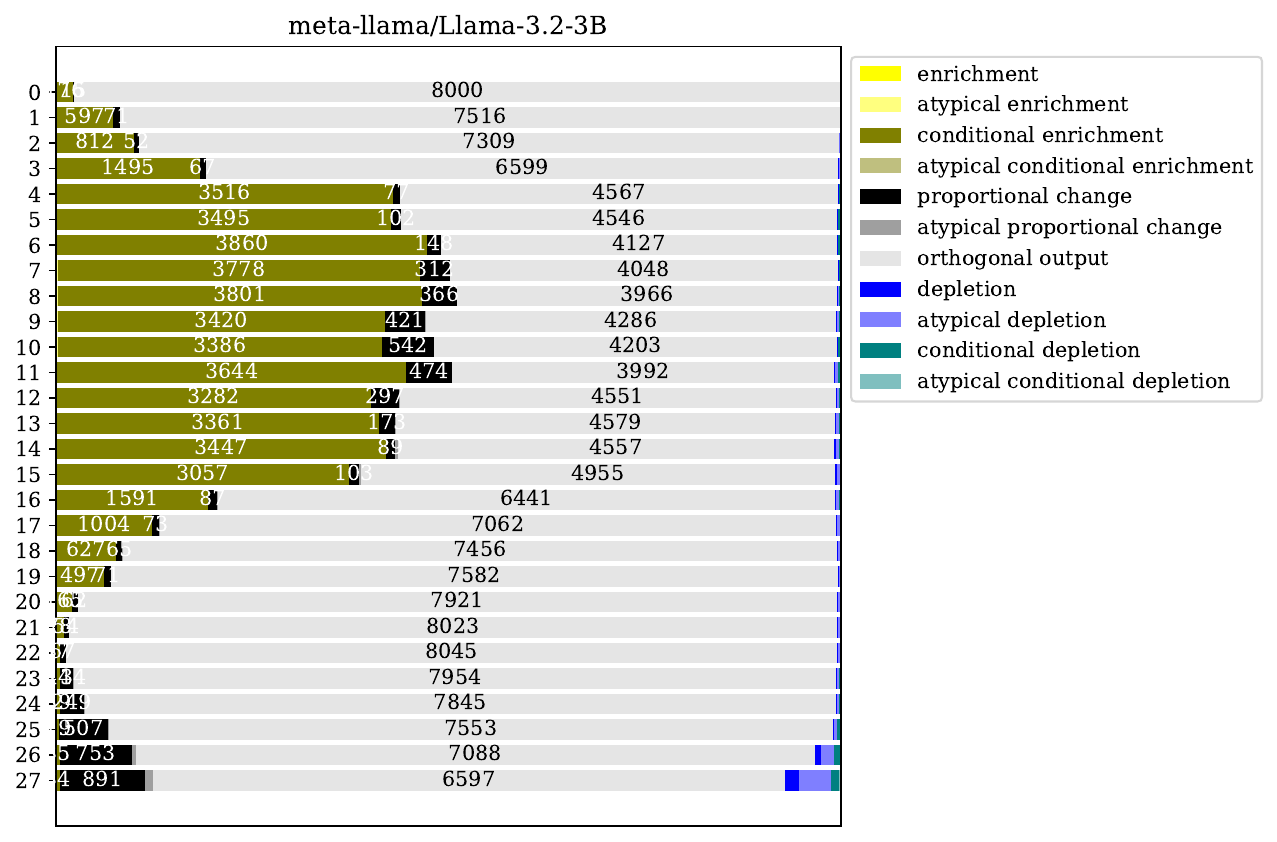}
	\caption{Distribution of neurons by layer and category. }
	\label{fig:bar}
\end{figure}

\cref{fig:bar} shows IO class distribution  across layers.

We see that a large proportion of neurons
are input manipulators
(i.e., they are not orthogonal output neurons):
in the Llama model,
these are 25\% of all neurons,
and as much as 50\% in early-middle layers
(layers 7--11).
This 
highlights an advantage of our
parameter-based IO classes:
It is an exhaustive analysis of all neurons,
and we can make non-trivial statements about a large subset of them.
Other methods 
only  assign a  subset of neurons to
classes; e.g.,
\citet{2024_Gurnee}'s classification only covers
1-5\% of neurons.

The majority of these input manipulators
(more than 80\% in Llama)
belong to just one class: conditional enrichment.
Across all models, conditional enrichment dominates early-middle layers.
In contrast, the (relatively few) input manipulators in late layers are often proportional change neurons or depletion neurons.

The dominance of
conditional enrichment neurons
in early-middle layers
echoes
\citet{geva-etal-2023-dissecting}'s and
\citet{lad2024remarkablerobustnessllmsstages}'s 
findings that these layers perform enrichment (or feature engineering).
We discuss this in \cref{sec:discussion-enrichment}.

These patterns hold for all models.
Some other models display additional patterns,
for example a large number of conditional depletion neurons in middle-late layers.
See \cref{ap:hr}.

\subsection{Marginal distributions}\label{sec:boxplots}

\begin{figure}
	\centering
	\includegraphics
	[width=\linewidth]
	{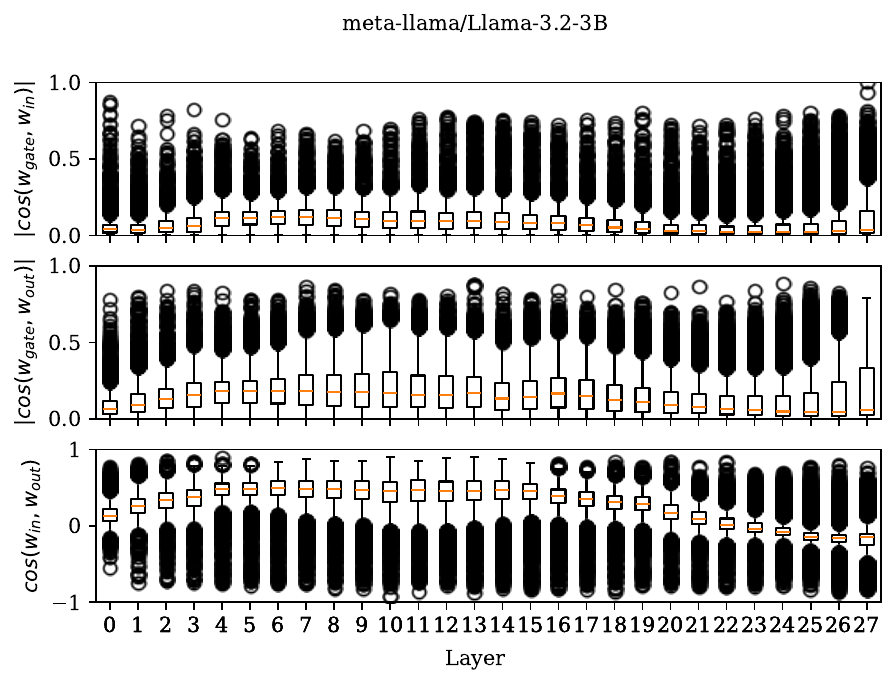}
	\caption{Boxplots for the distribution of weight cosine similarities in each layer.
	For $\cos(w_{\text{gate}},w_{\text{in}})$ and $\cos(w_{\text{gate}},w_{\text{out}})$ we show the absolute value since their sign does not carry any information on its own.}
	\label{fig:boxplots}
\end{figure}

\cref{fig:boxplots} shows the distribution of weight cosine similarities in each layer. 
In \cref{fig:medians} we also show
the median of $\cos(w_{\text{in}},w_{\text{out}})$,
across all investigated models.

We already know that
conditional enrichment neurons
are plentiful in the early-middle layers.
Correspondingly, the median value of $\cos(w_{\text{in}},w_{\text{out}})$
peaks in these layers.
Later on, it moves below zero, indicating that now the
majority of neurons have negative $\cos(w_{\text{in}},w_{\text{out}})$.
\cref{fig:medians} shows that this generalizes across models.

Regarding $|\cos(w_{\text{gate}},w_{\text{out}})|$,
the median values are relatively close to zero
(corresponding to conditional classes and orthogonal output).
But there is a large spread
in early-middle layers and in the last few layers.
This seems to correspond to the proportional change neurons
appearing in all of these layers,
as well as depletion neurons in the last few layers.

$|\cos(w_{\text{gate}},w_{\text{in}})|$ is mostly concentrated around zero.
Thus most neurons operate on two input directions in the
residual stream (not a single one), resulting in higher
expressivity and more complex semantics.
This is likely related to double checking;
see \cref{sec:doublecheck}.

We also notice that there are many outliers
for all three cosine similarities,
in almost all layers.
This suggests that a non-negligible number of neurons
perform special tasks different from the ``average'' neuron.

\enote{hs}{say that all the figures / tables are about
llama-3.2-3B somewhere? then you don't have to say it in the captions}

\subsection{Fine-grained analysis of IO behavior}\label{sec:finegrained}

\begin{figure}
	\centering
	\includegraphics
	[width=\linewidth]
	{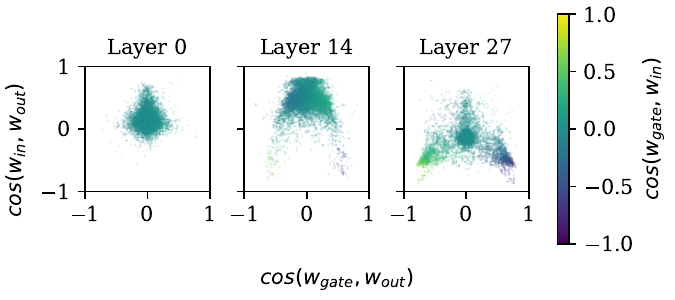}
	\caption{
	Fine-grained analysis of neuron IO
	behavior in three layers based on the configuration of
	their three weight vectors in parameter space.
Each subplot represents a layer, each dot
	a neuron.
	\label{fig:wcos_selected}}

\end{figure}

We now investigate weight vector configurations
in detail, as shown in
\cref{fig:wcos_selected} for
a few selected layers.
The distribution of neurons in each layer is
plotted
by displaying each neuron as a point with
$\cos(w_{\text{gate}},w_{\text{out}})$ indicated on the x-axis,
$\cos(w_{\text{in}},w_{\text{out}})$ on the y-axis
and
$\cos(w_{\text{gate}},w_{\text{in}})$ as its color.

This visualization reinforces three findings from \cref{sec:discrete,sec:boxplots}.
(i) We already know that many neurons are input manipulators.
Now we see that,
even though there are many neurons we classified as orthogonal output,
there is no cluster around the
origin as we might expect.
Instead, the orthogonal output neurons often
belong to clusters that are centered above/below
the horizontal line.
This suggests that even the orthogonal output neurons perform input manipulation to some extent.
(ii) 
We also have already observed
a smooth transition
from enrichment-like functionalities in early-middle layers
to more depletion-like functionalities in the last few layers.
We indeed see a large cluster of neurons,
centered clearly above the x-axis in most layers,
but moving below it in the last few layers.
(iii) We also observe that the vast majority of neurons is
turquoise, i.e.,
$\cos(w_{\text{gate}},w_{\text{in}}) \approx 0$,
confirming the finding in \cref{sec:boxplots}.

We also gain four new insights.
(i) 
The \emph{first layer} exhibits quite different patterns from model to model.
(ii)
In \emph{middle layers},
all models have a big cluster related to conditional enrichment neurons, as described above.
Additionally, many models
have outlier ``arms'' from this cluster,
towards the plot areas corresponding to proportional change and depletion.
Other models, such as OLMo, additionally have a cluster of neurons below the x-axis,
corresponding to conditional depletion neurons.
(iii)
Neurons with orthogonal $w_{\text{gate}}$ and $w_{\text{in}}$ must be within the unit disk.
It is striking to see that they do not fill out this disk evenly.
Instead, as already mentioned, there is a big cluster above the x-axis (close to conditional enrichment).
But this cluster is not right at the border of the disk, but more inside
(in particular $\cos(w_{\text{in}},w_{\text{out}})$ is still clearly below 1).
This echoes and extends
\citet{2024_Gurnee}'s findings
that in GPT2  the IO
cosine similarity is approximately bounded by $\pm
0.8$.
In other words, we almost never get the \textit{prototypical} cases
of conditional enrichment / depletion etc.,
as defined in \cref{sec:theory}.
This helps us refine our notion of ``input manipulators'':
these neurons do more than just
outputting a $w_{\text{out}}$ that is already present in the residual stream;
instead, they add novel but related information.
(iv)
In the \emph{last few layers} (Llama: layers 25-27), some new phenomena occur:
apart from  the big cluster,
there is a new cluster in the bottom corners of the plot (close to depletion).
Additionally, in the last layer of some models, 
there is a
cluster of turquoise points around the upper y-axis
(close to conditional enrichment).
\enote{sg}{it would be interesting to have a case study of one of these neurons}

\section{Case studies}\label{sec:cs}
We now demonstrate how the IO perspective can complement other methods
to help understand individual neurons.
To this effect,
we present 6 case studies for OLMo-7B, one for each
discrete
IO class.
We restrict the search space to
prediction/suppression neurons
(two of the output-based \textit{functional roles} of \citealp{2024_Gurnee}), 
i.e., each of the six neurons is a
prediction/suppression neuron as well as exemplifying one of our six classes.
For ease of interpretability, we choose that
prediction/suppression neuron
of a particular IO class with the highest
$\cos(w_{\text{out}}, W_U)$ kurtosis,
where $W_U \in \mathbb{R}^{d_{\text{model}}\times d_{\text{vocab}}}$ denotes the unembedding matrix.
(For orthogonal output we chose the clearest of all suppression neurons.)
The 6 neurons
are in the last layers of the model because that's where prediction/suppression neurons tend to appear.

See \cref{ap:functionalroles} for details
on prediction/suppression,
\cref{ap:cs-details} for more details on these case studies,
and \cref{ap:cs} for more case studies.

\subsection{Methods}
We combine the IO perspective with two well-established neuron analysis methods.
For each neuron, we project its weight vectors
to vocabulary space with the unembedding matrix $W_U$ and
inspect high-scoring tokens.
(This is analogous to \citep{2020_LogitLens}
and has been done e.g. in \citep{geva-etal-2022-transformer,2024_Gurnee,voita-etal-2024-neurons}.)
Additionally, we examine examples for which the neuron is
strongly activated (positive or negative)
among a subset of 20M tokens from Dolma \cite{soldaini-etal-2024-dolma}, OLMo's training set.
(Activation-based analyses have been done e.g. in \citealp{geva-etal-2021-transformer,voita-etal-2024-neurons,2024_Gurnee}.
The size of 20M tokens follows \citealp{voita-etal-2024-neurons}.)

\enote{sg}{``with the output embedding matrix''
Possibly also with the input embedding matrix, for reading weights that are not comparable to the output weight.
Not done yet
and not a priority.}






\subsection{Analysis}\label{sec:ss:res-cs}


For many of these neurons,
the largest positive activation is much larger than the largest negative one (or vice versa).
Often the larger of the two is also more interpretable.
In these cases we just describe the larger activation
and refer to \cref{tab:cs2} in \cref{ap:cs} for more details.

\textbf{Enrichment neuron 28.4737}
predicts \textit{review} (and related tokens) if
activated positively, which happens if \textit{review} is already
present in the residual stream. The maximally positive
activations are in standard contexts that continue with \textit{review} or similar,
such as
the newline after the description of an e-book
(the next paragraph often is the beginning of a review).

\textbf{Conditional enrichment neuron 28.9766's}
IO functionality concerns
\textit{well} and similar tokens.
28.9766
promotes them if activated positively, which happens
when both $w_{\text{gate}}$ and $w_{\text{in}}$ indicate that \emph{well} is
represented in the residual stream.
This is a case of double checking.
The maximally positive activation in our sample occurs on
\textit{\textbf{Oh}},
in a context in which \textit{Oh, well} makes sense
(and is the actual continuation).


\textbf{Depletion neuron 31.9634.}
$-w_{\text{out}}$ of 31.9634 is closest to forms of \textit{again}.
Judging by the weights,
the neuron activates positively when 
the residual stream contains
information both for and against predicting \textit{again},
and then depletes the \textit{again} direction.
It activates negatively when
the residual stream contains the ``minus \textit{again}'' direction,
and then depletes that direction.
Surprisingly, despite its strong negative cosine similarity
($\cos(w_{\text{gate}},w_{\text{in}})=-0.7164$), the neuron often
activates positively.
On the positive side,
strong activations are often on punctuation,
and the actual next token is often \textit{meanwhile} or \textit{instead}.
The neuron may ensure only these tokens are predicted,
and not the relatively similar \textit{again}.
On the negative side,
the activations do not have any obvious semantic relationship to \textit{again}.
We hypothesize that sometimes the residual stream ends up near ``minus \textit{again}''
for semantically unrelated reasons
(there are many more possible concepts than dimensions,
so the corresponding directions cannot be fully orthogonal;
see \citealp{2022_Elhage});
in these cases the neuron would reduce the unjustified presence of this ``minus \textit{again}'' direction.
There are also
weaker negative activations
when \textit{again} is a plausible continuation,
e.g., on the token \textit{\textbf{once}}.
In
these cases, \textit{again} is already weakly present in the
residual stream before the last MLP.  Accordingly,
$Swish(w_{\text{gate}} \cdot x_{\text{norm}})$ is weakly negative (but
distinct from 0), and $w_{\text{in}} \cdot x_{\text{norm}} > 0$, which leads
to a negative activation and thus reinforces \textit{again}.

\textbf{Conditional depletion neuron 29.10900.}
Gate and
linear input weight vectors act as two independent
ways of checking that \textit{these} is not present in the
residual stream (i.e., a case of double checking).
At the same time, they check for predictions like \textit{today, nowadays}.
When such predictions are present,
the neuron
promotes \textit{these}.
This is a plausible choice in these cases because of the expression \textit{these days}.
An example is
\textit{social media tools change and come and go at the drop of a \textbf{hat}}.
(This sentence talks about a characteristic of current times,
so \textit{these days} would indeed be a plausible continuation.)

\textbf{Proportional change neuron 30.10972}
predicts
the token \textit{when} if activated negatively. This happens if
\textit{when} is absent from the residual stream (gate
condition) and is proportional to the presence of time-related
tokens (-$w_{\text{in}}$).
An example for a large negative activation is
\textit{puts you on multiple webpages \textbf{at}}.\footnote{
The actual sentence ends with \textit{as soon as} and comes from a now-dead webpage.
We also found one occurrence of \textit{at when} in what seems to be a paraphrase of the same text,
on https://www.docdroid.net/RgxdG5s/fantastic-tips-for-bloggers-of-all-amountsoystcpdf-pdf .
We suspect that both texts are machine-generated paraphrases of an original text containing \textit{at once}
(\textit{when} and \textit{as soon as} can be synonyms of \textit{once} in other contexts),
and that the model has (also) seen a paraphrased version with \textit{at when}.
In fact many of the largest negative activations are on \textit{at} in contexts calling for \textit{at once}.
}
Conversely, if \textit{when} is absent, and time-related tokens are absent too,
the neuron activates positively and suppresses \textit{when} further.

\textbf{Orthogonal output neuron
29.4180} predicts \textit{there} (positive activation) if
the residual stream contains a component that we
interpret as ``complement of place expected''
(e.g., \textit{here}, \textit{therein}).
Both
$w_{\text{gate}}$ and $w_{\text{in}}$ check for (different aspects of)
this component being present, another case of
double checking.
The largest positive activation is on
\textit{here \textbf{or}}.

Overall, these neurons all promote a specific set of tokens
(we chose them that way),
but under very different circumstances.
The (conditional) enrichment neurons
are the most straightforward to interpret,
because their input and output clearly correspond to the same concept.
In contrast, depletion neurons inherently involve
(an apparent) conflict
between the intermediate model prediction and what the neuron promotes.

\section{Discussion}\label{sec:discussion}

\enote{hs}{do we need this? it's clear from the subsection
headings what the discussion topics are
We now discuss the implications of our findings.
We start with general observations,
on variation across models and on double checking.
Then we focus on the IO classes
(conditional) enrichment and (conditional) depletion.
}

\enote{sg}{Do we need these two?:
\shortpar{Input manipulation}
Many
neurons
perform input manipulation;
the fine-grained analysis of \cref{sec:finegrained} shows this especially well.
However, they do so in various ways, and not all of the
theoretical possibilities are implemented
equally.
Conditional enrichment (especially in early-middle
layers) seems to be the most useful
pattern for the model.
Other relevant patterns are
depletion (especially in the last few layers),
and, for some models (in middle-late layers), conditional depletion.
Proportional change neurons (acting on the gate direction only)
also play a role,
but they do not form clear clusters in \cref{fig:wcos_selected}.
The final major IO
functionality, enrichment of a single input direction
(enrichment neurons), is rare.

\shortpar{Functional roles}
We restricted the neurons in our case studies (\cref{sec:cs})
to those assigned by \citet{2024_Gurnee} to the
functional role prediction/suppression.
Our classification sheds light on important differences
between neurons assigned to the same functional role,
showing that our classification of
IO behavior
is an important complement.
}


\subsection{Variation across models}
Our work on gated activation functions
questions the generality of previous findings \cite{voita-etal-2024-neurons,2024_Gurnee}
on non-gated activation functions.
Specifically, we saw in \cref{sec:exp} that
(conditional) depletion neurons appear mostly in later layers.
On the other hand,
\citet{2024_Gurnee}
find
(for GPT-2 \cite{radford2019language}, with activation GeLU)
that
what we call depletion neurons
mostly appear in earlier layers.
Similarly, \citet{voita-etal-2024-neurons}
find
(for OPT \cite{zhang2022opt}, with activation ReLU)
that some neurons in early layers detect specific tokens and
then suppress them.
(Their analysis is not weight-based,
so these may or may not be depletion neurons in our weight-based sense.)

This confirms the importance of our work for models with gated activation functions:
their internal structure is quite different from older models with GeLU or ReLU.

Despite minor differences (especially in the first layer), 
our results 
across gated activation models are
remarkably consistent. Most importantly, all of them are
dominated by conditional enrichment neurons in early-middle
layers and all of them tend
towards depletion in the very last layers.

\enote{hs}{cut for space
However, there are still some differences between
these models, especially regarding the first layer, and the
presence of conditional depletion neurons.  These
differences are not obviously correlated to any differences
in architecture, so future work is needed to explain them.
Possibly the various models have just found different local
minima in loss space.  In any case, these minor differences
cannot overshadow the broad agreement across models.
}

\subsection{Double checking}\label{sec:doublecheck}
Our case studies suggest that
conditional enrichment or conditional depletion neurons often behave in a way analogous to their unconditional counterparts.
One reason is
that our threshold for distinguishing 
conditional and unconditional classes is somewhat
arbitrary.

These and other neurons
(for example, proportional change neurons like 25.8607, \cref{ap:cs})
display a phenomenon we called double checking:
They use two quite different reading weight vectors to check for a single concept.


Double checking is rooted in the following geometric fact:
Two vectors $w_1, w_2$ ($w_{\text{gate}}$ and $w_{\text{in}}$ in our case) can be orthogonal to each other but still have a high similarity to a third vector $u$ (e.g., a token unembedding).
Example: $w_1 = (1,0), w_2=(0,1), u = (1,1)$.
Here, $w_1,w_2$ are orthogonal, but $u$ has a cosine of $\frac{\sqrt{2}}{2}\approx 0.7$ to both.

Double checking is useful because it shrinks the region in model space that activates the neuron positively.
If (say) $w_{\text{in}}=w_{\text{gate}}= (1,0)$, the neuron activates whenever the (normalized) residual input $x$ satisfies $x \cdot (1,0) > 0$;
this happens on the whole half-space $x_1>0$.
If however $w_{\text{gate}} = (1,0)$ and $w_{\text{in}} = (0,1)$, the neuron activates positively only in the first quadrant ($x_1,x_2>0$).

This behavior thus enables more precise concept detection.
This may explain
why conditional neurons are more frequent than their unconditional counterparts.

\subsection{Stages of inference}
\label{sec:stages}
\enote{sg}{
I'm commenting out the experiments, they absolutely don't convince me anymore.
I'm trying to do convincing ablation experiments, but it's unlikely I'll be finished before the deadline.
}

We saw in \cref{sec:exp} that
different layers are dominated by different IO functionalities.
This leads to a follow-up question:
Why does the model use these specific IO functionalities in these specific layers?
In particular:
Why are there so many conditional enrichment neurons in early-middle layers?
And what is the role of (conditional) depletion neurons in later layers?
We hypothesize
that different IO classes might be responsible for different
\textit{stages of inference} \cite{lad2024remarkablerobustnessllmsstages},
as described in the following subsections.
In future work, we plan to test this hypothesis using ablation experiments.

\enote{hs}{above: should incorprate out discussion from Friday}

\subsection{Enrichment}\label{sec:discussion-enrichment}
We saw in \cref{sec:exp} that there often is
positive similarity between
reading
and writing
weights of neurons,
especially with conditional enrichment neurons in early-middle layers.

These neurons
seem a good fit
for the \textit{feature engineering} stage
\citep{lad2024remarkablerobustnessllmsstages},
corresponding to enrichment as defined by \citet{geva-etal-2023-dissecting}.
Indeed, they
output a direction similar to the one they detect,
which could correspond to related concepts.
\citeposs{geva-etal-2023-dissecting} description of enrichment precisely involves writing related concepts to the residual stream.

In later layers, the (conditional) enrichment neurons we investigated
in our case studies (\cref{sec:cs})
have an output that is semantically identical to the input.
Thus they seem to reinforce existing predictions.

In general,
we use the
term \emph{enrichment} because the output weight is never
mathematically identical
to one of the reading weights.
But depending on the
analysis of a particular neuron
(e.g., by way of a case study),
magnification (no
change)
or enrichment (e.g., change 
\textit{Ireland} in the input to 
\textit{Dublin} in the output)
may be the more intuitive
human interpretation.


\subsection{Depletion}\label{sec:depletion}
We saw in \cref{sec:exp} that
depletion neurons appear mostly in the last few layers,
and conditional depletion neurons appear in later-middle layers (if at all).

These neurons
reduce the presence of the directions they detect.
Therefore they seem a good fit for the \textit{residual sharpening} stage
-- getting rid of attributes that are not directly needed for next token prediction.



We found depletion
neurons more difficult to interpret than enrichment neurons. Most notably, neuron 
31.9634 was a complex case
in that we found contexts in which a weak positive presence 
of \emph{again}
led to an
enrichment-like functionality (see \cref{sec:ss:res-cs}).
This mechanism involves a negative value of Swish.
Previous authors \cite{2023_Gurnee} often assumed that GELU (or equivalently, Swish) is
``essentially the same activation as a ReLU'',
and said they 
``would be particularly excited to see future work exhibiting [...] case studies''
of mechanisms involving negative values of such an activation function.
To our knowledge,
we show for the first time
that negative values of Swish can play a crucial role in
how transformers function.

Still, all neurons we investigated do deplete input
directions from the output even if they do not do so in all contexts. 
We plan to further elucidate
the intuitive role depletion plays in follow-up work.



%




\section{Conclusion}

We  explored the IO perspective for investigating gated neurons in LLMs.
Our method complements prior interpretability approaches
and provides new insights into the inner workings of LLMs.

We observed that a large share of neurons exhibit non-trivial IO interactions.
The concrete IO functionalities differ from layer to layer,
which is probably related to different stages of inference.
In particular, early-middle layers are dominated by conditional enrichment neurons,
which may be responsible for representation enrichment.

We plan to further develop this new perspective in future work.
In particular, we will do ablation experiments to conclusively show
if, as we hypothesized,
the conditional enrichment neurons in early-middle layers are responsible for representation enrichment
and the depletion neurons in the last few layers contribute to residual sharpening.
We also plan to investigate the evolution of IO functionalities during model training.
Finally, we would like to go beyond the analysis of single neurons
and address the question of how neurons work together within and across IO classes.

\section*{Limitations}
This paper focuses on a \textit{parameter-based} interpretation of \textit{single neurons}.
This has the advantage of being simple and efficient,
but is also inherently limited in scope.
Accordingly, our method is not designed to replace other neuron analysis methods,
but to complement them.

The mathematical similarities of weights are insightful,
but they should not be taken as one-to-one representations of semantic similarity:
We find cases in which close-to-orthogonal vectors represent very similar concepts (\cref{sec:doublecheck}),
and cases in which mathematically similar vectors represent related but non-identical concepts (\cref{sec:stages}).

Our case studies of individual neurons can be accused of cherry-picking:
we picked neurons that we expected to be interpretable,
all of which occur on the last few layers.
Therefore our interpretations may not carry over to less interpretable (e.g. polysemantic) neurons,
or to neurons in earlier layers.

Finally, we provide only possible interpretations of the phenomena we observe, and do not claim them to be definitive explanations.

\bibliography{custom,anthology}

\appendix

\section{Overview of the appendix}

\cref{ap:sd}: Software and data.

\cref{ap:impact}: Impact statement.

\cref{ap:responsible}: ``Responsible NLP'' statements.

\cref{ap:swiglu}:
Visualization of a SwiGLU neuron (\cref{sec:ss:swiglu}).

\cref{ap:functionalroles}: IO classes vs. \citeposs{2024_Gurnee} \textit{functional roles}. Used in \cref{sec:cs}.

\cref{ap:cs-details}: Details on case studies (\cref{sec:cs}).

%

\cref{ap:cs}: More case studies (complementing \cref{sec:cs}).

\cref{ap:hr}: Results across models (complementing \cref{sec:exp}).

We chose to put the last section at the end
because it is very long and would otherwise disrupt reading of the other sections.

\section{Software and data}\label{ap:sd}
\enote{For the review version:
This review version is accompanied by zip archives containing software and data, respectively.
See the readme file for detailed documentation.

We plan to release the software under a permissive license such as Apache 2.0.

The data archive currently contains only
the visualizations of max/min activations for the neuron case studies in \cref{sec:cs}.
Everything else can be quickly reproduced,
and the plots are included in this paper.
We plan to release these visualizations under the Apache 2.0 license
(they contain text from Dolma, which is under the same license).
}

We publish the software at \href{https://github.com/sjgerstner/IO_functionalities}{this Github URL}.
See the readme file for detailed documentation.

The repository also contains
the visualizations of max/min activations for the neuron case studies in \cref{sec:cs}.
Everything else can be quickly reproduced,
and the plots are included in this paper.

The repository is under the Apache 2.0 license.

\section{Impact statement}\label{ap:impact}
This paper presents work whose goal is to advance the field
of machine learning interpretability.  The underlying
assumption of the field is that models have underlying
structure (are not just an inscrutable mess) and that
discovering this structure will have several benefits.
First, ideally, any scientific field should have a deep
understanding of the models it uses; results that are
obtained using blackbox models are hard to understand,
replicate and generalize. Second, once we understand our
models better, we will be better able to address failure
modes. For example, once we understand how unaligned
behavior like bias and hallucinations comes about, it will
be easier to address them, e.g., by changing the model
architecture.  Third, interpretability can support
explainability. If we understand how a recommendation or
answer came about, we can better assess its validity.

\section{``Responsible NLP'' statements}\label{ap:responsible}
\subsection{Models and data}

\shortpar{Gemma}
To download the model one needs to explicitly accept the
\href{https://ai.google.dev/gemma/terms}{terms of use}.
NLP research is explicitly listed as an intended usage.
Primarily English and code
\cite{gemma_2024}.

\shortpar{Llama}
Inference code and weights under an ad hoc
\href{https://github.com/meta-llama/llama/blob/main/LICENSE}{license}.
There is also an
\href{https://github.com/meta-llama/llama/blob/main/USE_POLICY.md}{``Acceptable Use Policy''}.
Our work is well within those terms.
Languages mostly include English and programming languages,
but also Wikipedia dumps from
``bg, ca, cs, da, de, en, es, fr, hr, hu, it,
nl, pl, pt, ro, ru, sl, sr, sv, uk''
\cite{llama}.

\shortpar{OLMo and Dolma}
Training and inference code, weights (OLMo), and data (Dolma) under Apache 2.0 license.
``The Science of Language Models'' is explicitly mentioned as an intended use case.
Dolma is quality-filtered and designed to contain only English and programming languages
(though we came across some French sentences as well, see \cref{tab:cs2})
\cite{groeneveld-etal-2024-olmo,soldaini-etal-2024-dolma}.

\shortpar{Mistral}
Inference code and weights are released under the Apache 2.0 license,
but accessing them requires accepting the
\href{https://mistral.ai/terms}{terms}.
Languages are not explicitly mentioned in the paper,
but clearly include English and code
\cite{jiang2023mistral7b}.

\shortpar{Qwen}
Inference code and weights under Apache 2.0 license.
Supports ``over 29 languages, including
Chinese, English, French, Spanish, Portuguese, German, Italian, Russian, Japanese, Korean, Vietnamese, Thai, Arabic,
and more''
\cite{qwen2}.

\shortpar{Yi}
Inference code and weights under Apache 2.0 license.
Trained on English and Chinese
\cite{ai2025yiopenfoundationmodels}.

\subsection{Computational experiments}
All our experiments can be run on a single NVIDIA RTX A6000 (48GB).
The main analysis, computing the weight cosines, needs less than a minute per model.
The most expensive part was the activation-based analysis in \cref{sec:cs}:
We needed a single run of $\approx 25$ h to store the max/min activating examples for all neurons,
and then $\approx 45$ s per neuron ($\approx 5$ min) to recompute its activations on the relevant texts and visualize them.

We use TransformerLens \cite{nanda2022transformerlens}.
A colleague kindly provided us with
\href{https://anonymous.4open.science/r/TransformerLens-0EA4/}{a version}
that also supports OLMo.

\begin{figure*}
	\centering
	\begin{tikzpicture}[
	scale=1,
parameter/.style={rectangle, draw=blue!60, fill=blue!5, very thick, minimum size=7mm},
state/.style={rectangle, draw=black, very thick, minimum width=15mm, minimum height=7mm},
scalar/.style={ellipse, draw=black, very thick, minimum width=15mm},
other/.style={rectangle, draw=black, very thick, minimum width=7cm},
]
\node[state] (xln)
{$x_{\text{norm}}$};
\node (empty) [above right=1cm of xln] {};
\node[scalar] (xin) [right=1cm of empty] {$x_{\text{in}}$};
\node[scalar] (xgate) [below right=1cm of xln] {$x_{\text{gate}}$};
\node[scalar] (swish) [right=1.25cm of xgate] {$x_{swish}$};
\node[scalar] (f) [right=5cm of xln] {$f(x_{\text{norm}})$};
\node[state] (xout) [right=1.25cm of f] {$x_{\text{out}}$};

\draw[->] (xln.south east) -- node[parameter, below left=1mm]{$\cdot w_{\text{gate}}$} (xgate.north west);
\draw[->] (xln.north east) -- node[parameter, above left=1mm]{$\cdot w_{\text{in}}$} (xin.west);
\draw[->] (xgate.east) -- node[below] {Swish} (swish.west);
\draw[->] (swish.north east) -- node[above left]{multiply} (f.south west);
\draw[->] (xin.east) --  (f.north west);
\draw[->] (f.east) -- node[parameter, above=1mm]{$\cdot w_{\text{out}}$} (xout.west);
\end{tikzpicture}
	\caption{Visualization of the SwiGLU activation function for a single neuron. Boxes represent vectors, ellipses represent scalars.}
	\label{fig:swiglu}
\end{figure*}
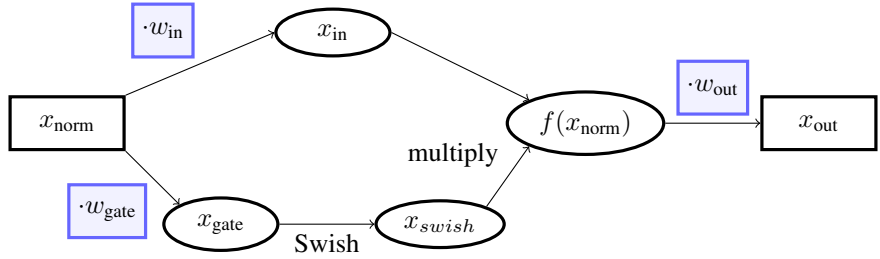

\section{More on SwiGLU}\label{ap:swiglu}
\cref{fig:swiglu} visualizes a SwiGLU neuron (described in \cref{sec:ss:swiglu}).
%

\section{IO classes vs. functional roles}\label{ap:functionalroles}

We compare our results with those of another classification scheme we mentioned in \cref{sec:ss:related}: the \emph{functional roles} defined by \citet{2024_Gurnee}. See \cref{sec:ss:res-matrix} for the results.

\subsection{Definition of functional roles}\label{ap:def-functionalroles}
The definition of functional roles is based exclusively on the neuron's output weight $w_{\text{out}}$.
Most of the roles are defined by
their output token distribution, i.e., properties
of the distribution $\cos(w_{\text{out}},W_U)
= \left(\frac{w_{\text{out}}\cdot
W_U[:,1]}{\|w_{\text{out}}\|\|W_U[:,1]\|}, ..., \frac{w_{\text{out}}\cdot
W_U[:,d_{\text{vocab}}]}{\|w_{\text{out}}\|\|W_U[:,d_{\text{vocab}}]\|} \right) \in
[-1,1]^{d_{\text{vocab}}}$, the cosine of the product of output
weight vector and unembedding matrix.


Functional roles are defined as follows.
	\textbf{Prediction} and \textbf{suppression} neurons
	have a
	$\cos(w_{\text{out}},W_U)$ with
	high \textit{kurtosis} (meaning
	there are many outliers) and a high
	\textit{skew} in absolute value (meaning the outliers tend to be only on one
	side). Positive skew corresponds to predicting a 
	subset of tokens, negative skew to suppressing
	it.
	\textbf{Partition} neurons have a distribution $\cos(w_{\text{out}},W_U)$ with high \textit{variance}.
	This often corresponds to two sets of output tokens, one
	that is promoted and one that is suppressed.
	In \textbf{entropy} neurons
	(examined in more detail by \citet{2024_Stolfo},
	$w_{\text{out}}$ lies in a direction that does not correspond to any output tokens.
	Mathematically,
	a high proportion of
	the norm of $w_{\text{out}}$ is in $W_U$'s effective null space,
	i.e., it corresponds to singular vectors of $W_U$ whose corresponding singular values are close to zero.
	Entropy neurons
	increase or decrease
	the presence of such directions.
	This changes the norm of the
	residual stream, but leaves the token ranking more or less
	untouched. Because a final LayerNorm is applied before
	$W_U$, this indirectly affects the logits of all tokens:
	the output token probabilities become more evenly distributed (higher entropy), or less so (lower entropy).
	 \textbf{Attention (de)activation} neurons
	(de)activate an
	attention head by having it  put less (or more) of its attention
	on the BOS token. (The effect of a head attending only to
	BOS is negligible.)
	Consider an attention head with query matrix
	$W_Q \in \mathbb{R}^{d_{\text{model}}\times d_{head}}
	= \mathbb{R}^{4096\times 128}$
and
	BOS key vector $k_{BOS}\in\mathbb{R}^{d_{head}}$. Attention
	(de)activation neurons for this head are those with a high
	positive or negative score $w_{\text{out}}W_Q
	k_{BOS}$. 

All of these definitions require a threshold and/or some adaptation to gated activation functions.
We describe our approach in \cref{sec:thresholds}.

\subsection{Adapting the definitions}\label{sec:thresholds}
The \textit{functional role} definitions require a threshold and/or some adaptation to gated activation functions.
We proceed as follows:
\begin{itemize}
	\item We set the number of \textit{partition} neurons to be 1000, which gives a variance of 0.0007 as a threshold.
	\item Preliminary
	experiments show that (absolute) skew and kurtosis are
	highly correlated in practice, so we decide to focus on
	kurtosis to find prediction / suppression neurons.
	We then choose a kurtosis threshold for \textit{prediction/suppression}, such that the \textit{prediction/suppression} class is disjoint from \textit{partition}. This gives a (very high) excess kurtosis of 230.9736.
	\item  \textit{Entropy}: Following \citet{2024_Stolfo}, we focus on the last layer, and we define the null space of $W_U$ as the subspace of model space spanned by its last 40 singular vectors. We find that two neurons have a particularly high proportion of their norm in this null space, and define these as entropy neurons.
	\item \textit{Attention (de)activation}: To ensure comparability across heads, we normalize $w_{\text{out}}$ and $W_Q k_{BOS}$. Thus the scores can be intuitively understood as cosine similarities between these two vectors. We choose $\pm \frac{\sqrt{2}}{2}$ as a cutoff. We keep only those neurons that we did not already classify as partition or prediction/suppression.
	\item In our case the neuron can be activated
	positively or negatively, so we cannot distinguish
	prediction from suppression \textit{a priori}. Instead, we automatically distinguish \textit{prediction} and \textit{suppression} from each other by the sign of $\cos(w_{\text{in}},w_{\text{gate}}) \cdot \mbox{skew}(\cos(w_{\text{out}},W_U))$ (as opposed to just the sign of the skew). The quantity $\cos(w_{\text{in}},w_{\text{gate}})$ indicates the typical sign of the activation \textit{a priori}. Even though this is not very trustworthy it gives some interesting results. 
	\item The same problem occurs for the distinction of attention activation and deactivation. As before, we multiply the original quantity
	$w_{\text{out}} W_Q k_{BOS}$
	by $\cos(w_{\text{in}},w_{\text{gate}})$ and only then look at the sign.
	Note that here a positive sign means high attention on BOS, hence attention \emph{de}activation.
	It turns out that all relevant neurons are attention \emph{de}activation according to this metric.
\end{itemize}

\subsection{Results}\label{sec:ss:res-matrix}
\begin{table*}\centering
\small

    \rowcolors{2}{white}{gray!25}
    \begin{tabular}{l|rrrrrrr|r}
    	&  &  &  &  & \multicolumn{2}{c}{attention}  &  & \\
    	& prediction & suppression & partition & entropy &  \multicolumn{2}{c}{deactivation} & other & total\\
    	\hline
    	depletion & 73 & 0 & 51 & 0 & 2 & 14 & 117 &243\\
    	at.\ depletion & 114 & 0 & 61 & 0 & 0 & 3& 429 &604\\
    	c.\ depletion & 68 & 1 & 24 & 2 & 0& & 12,344 &12,439\\
    	at.\  c.\  depletion & 19 & 0 & 13 & 0 &0 &  & 12 &44\\
    	orthogonal output & 826 & 203 & 516 & 0 &0& &190,832 &192,377\\
    	proportional change & 111 & 206 & 139 & 0 &0& 1& 23,358 &23,814\\
    	at.\  proportional change & 25 & 0 & 16 & 0 & 2 & & 85 &128\\
    	c.\  enrichment & 48 & 0 & 179 & 0 &0 & 1& 121,446 &121,673\\
    	at.\  c.\  enrichment & 14 & 0 & 0 & 0 & 0& & 660 &674\\
    	enrichment & 6 & 0 & 0 & 0 & 0&& 18 &24\\
    	at.\  enrichment & 15 & 0 & 1 & 0 & 0&& 220 &236\\
    	\hline
    	total & 1,319 & 410 & 1,000 & 2 & 4 &21 & 349,521 & 352,256
    \end{tabular}
	\caption{Contingency table of
    IO classes (rows) vs
\citet{2024_Gurnee}'s functional roles (columns) for OLMo-7B-0424.
c = conditional. at = atypical.
Cutoffs for prediction/suppression and partition were chosen
as described in \cref{sec:thresholds}.
Many neurons with high attention deactivation score are also
partition neurons; the left column unter ``attention
deactivation'' counts
only those that are not.
OLMo-7B-0424 has no attention activation neurons with high
enough score.
}
\label{tab:functional}
\end{table*}

The contingency matrix in \cref{tab:functional} is a
systematic comparison of our IO classes
with \citet{2024_Gurnee}'s
functional roles.

We first see again that \citet{2024_Gurnee} assign a
functional role to only a small proportion of all
neurons. 349,521 of 352,256 neurons remain unclassified. In
contrast, our IO classes are
exhaustive and robustly identify functionalities 
like conditional depletion and enrichment that are
explanatory for how transformers process language.

We find that 
prediction neurons, suppression neurons and (less consistently)
partition neurons mostly occur in the final layers, 
replicating \citet{2024_Gurnee}'s findings.
\enote{sg}{we need a plot or table to show this explicitly}

Most of these neurons are orthogonal output or proportional change.
This is not unexpected, as these are some of the largest classes.
Conversely, however, a majority of the (relatively few) depletion neurons have prediction or partition as functional role.
\enote{sg}{
We need to show this explicitly but I'm pretty sure it's true:
This remains the case if we account for the fact that both prediction / partition and depletion neurons mostly occur in the last few layers.
}

The only two entropy neurons in OLMo-7B-0424
occur in the last layer and are conditional depletion neurons.

\begin{table*}\centering
   \rowcolors{2}{white}{gray!25}
    \begin{tabular}{l|l|rrr}
    	Neuron & IO category & $\cos(w_{\text{gate}},w_{\text{in}})$ & $\cos(w_{\text{gate}},w_{\text{out}})$ & $\cos(w_{\text{in}},w_{\text{out}})$ \\
    	28.4737 & enrichment  &0.5290&0.5048&0.7060\\
    	28.9766 & conditional enrichment  &0.4764&0.4119&0.5982\\ 
   	31.9634 & depletion  &-0.7164&0.7218&-0.8542\\
    	29.10900 & conditional depletion &0.4988&-0.4992&-0.5775\\
    	30.10972 & proportional change & -0.4543&0.5814&-0.4182\\
    	29.4180 & orthogonal output &-0.0272&-0.4057&0.0669\\
    \end{tabular}
    \caption{Overview of prediction/suppression neurons chosen for case studies in \cref{sec:cs}}
    \label{tab:cs}
 \end{table*}

\begin{table*}\centering
    \rowcolors{2}{white}{gray!25}
    \begin{tabular}{p{5em}|p{4em}p{4em}|p{4em}p{4em}|p{4em}|p{15em}}
    
    	Neuron,\newline IO class & \multicolumn{2}{l|}{$w_{\text{gate}}$} & \multicolumn{2}{l|}{$w_{\text{in}}$} & $w_{\text{out}}$ & Top activations\\
    	
    	28.4737\newline enrichment &
    	\multicolumn{2}{l|}{$\approx w_{\text{out}}$} &
    	\multicolumn{2}{l|}{$\approx w_{\text{out}}$} &
    	pos:\newline \textit{\hspace*{.5em}review\newline \hspace*{.5em}Review}
    	&pos 
    	(13.75): \textit{Download EBOOK [...] Description of the book [...] \textbf{\textbackslash n} -> Reviews}
    	\newline
    	neg 
    	(-2.25): \textit{The answer's at the bottom of \textbf{this} -> post}
    	\\
    	
    	28.9766\newline conditional enrichment &
    	pos:\newline \textit{well\newline \hspace*{.5em}well
    	} & neg:\newline \textit{\hspace*{.5em}far\newline \hspace*{.5em}high
    	} &
    	\multicolumn{2}{l|}{$\approx w_{\text{out}}$} &
    	pos:\newline \textit{\hspace*{.5em}well\newline well
    	}
    	&pos 
    	(18.63): \textit{Could have saved myself some time. \textbf{Oh} -> , well}
    	\newline
    	neg 
    	(-3.66): \textit{Seek to understand them \textbf{more} -> fully}
    	\\
    	
    	31.9634\newline depletion &
    	\multicolumn{2}{l|}{$\approx w_{\text{out}}$} &
    	\multicolumn{2}{l|}{$\approx-w_{\text{out}}$} &
    	neg:\newline \textit{\hspace*{.5em}again\newline \hspace*{.5em}Again
    	}
    	&pos 
    	(5.12): \textit{jumping off the roof of his Los Angeles apartment building\textbf{.} -> Meanwhile}
    	\newline
    	neg 
    	(-3.48): \textit{the areas of the doorjamb where the \textbf{door} -> often}
    	\\
    	
    	29.10900\newline conditional depletion &
    	pos:\newline \textit{\hspace*{.5em}today\newline \hspace*{.5em}nowadays
    	}& neg:\newline \textit{\hspace*{.5em}these\newline these
    	} &
    	\multicolumn{2}{l|}{$\approx-w_{\text{out}}$} &
    	pos:\newline \textit{\hspace*{.5em}these\newline \hspace*{.5em}These
    	}
    	&pos 
    	(12.79): \textit{social media tools change and come and go at the drop of a \textbf{hat} -> .}
    	\newline
    	neg 
    	(-2.18): \textit{la couleur de sa robe \textbf{et} -> le}
    	\\
    	
    	30.10972\newline proportional change &
    	\multicolumn{2}{l|}{$\approx w_{\text{out}}$} &
    	pos:\newline \textit{\hspace*{.5em}when\newline when
    	}& neg:\newline \textit{\hspace*{.5em}timing\newline \hspace*{.5em}dates
    	} &
    	neg:\newline \textit{\hspace*{.5em}when\newline when
    	}
    	&pos 
    	(2.67): \textit{Take pleasure in the rest of the new year\textbf{.} -> You}
    	\newline
    	neg 
    	(-6.14): \textit{puts you on multiple webpages \textbf{at} -> as soon as}
    	\\
    	
    	29.4180\newline  orthogonal output&
    	pos:\newline \textit{\hspace*{.5em}here\newline \hspace*{.5em}therein
    	}& neg:\newline \textit{\hspace*{.5em}there\newline \hspace*{.5em}we
    	} &
    	pos:\newline \textit{\hspace*{.5em}here\newline \hspace*{.5em}in
    	}& neg: ? &
    	pos:\newline \textit{\hspace*{.5em}there\newline there
    	}
    	&pos 
    	(14.41): \textit{here \textbf{or} -> there}
    	\newline
    	neg 
    	(-2.31): \textit{without any consideration being issued or paid \textbf{there} -> for}
    	\\

    \end{tabular}
    \caption{Description of the weight vectors of the selected neurons, by top tokens or similarity to $w_{\text{out}}$.
    The question mark, ?, signals unknown unicode characters.
    The last column presents the (shortened) text samples on which the respective neuron activates most strongly (positively or negatively).}
    \label{tab:cs2}
\end{table*}

\section{Details on case studies}
\label{ap:cs-details}
See \cref{tab:cs,tab:cs2} for more details on the case studies of \cref{sec:cs}.

\enote{sg}{What to do about unicode in \cref{tab:cs2}?}

\section{More case studies}\label{ap:cs}\label{ap:res-cs}
These are various neurons that popped out to us as possibly interesting,
for not very systematic reasons,
for example because they strongly activated on a specific named entity.
All of them are in OLMo-7B.
We present them by IO class.
For most of these case studies we did only a quick and dirty weight-based analysis.
In some cases we also tried $W_E$ (input embeddings) instead of $W_U$ (unembeddings) for the logit-lens style analysis.

\enote{sg}{In a future version we could include
the case change neuron (conditional depletion, perhaps partition)
and/or the policy neuron (depletion, perhaps suppression)}

\subsection{Conditional enrichment neurons}
	
\textbf{0.1480}: $w_{\text{gate}}, -w_{\text{in}}, -w_{\text{out}}$ all have tokens similar to \textit{box} (when using $W_E$). Activates on \textit{Xbox}.

\textbf{4.1940}: \textit{country} appears in $w_{\text{in}}$ among many other things.
When using $W_E$, \textit{Philippines} and \textit{Manila} appear in $w_{\text{out}}$.
	Activates on \textit{Philippines}.

\textbf{4.3720}: gate seems country/government related.
	When using $W_E$, we find $w_{\text{out}}, w_{\text{gate}}$ contain some country names.
	Activates on \textit{Denmark}.

\textbf{4.4801}: \textit{Muhammad} appears in the gate vector.
Activates on \textit{Muhammad}.
	
\textbf{4.5772}: predicts \textit{ian} as in \textit{Egyptian}.
	When using $W_E$, all three weight vectors contain \textit{Egypt}.
	Activates on \textit{Egypt}.

\textbf{4.6517} has a very Ireland (or Celtic nations) related gate vector. The interpretations of the other two weights are less obvious, but \textit{Irish} and \textit{Dublin} appear in $w_{\text{in}}$ among many other things, and \textit{UK} and \textit{London} appear in $-w_{\text{out}}$ (Ireland is emphatically \textit{not} in the UK!)
When using $W_E$, \textit{Ireland} appears among the top tokens of all three weight vectors.
Activates on \textit{Ireland}.

\textbf{4.6799}: When using $W_E$, \textit{Vietnam} is among the tokens corresponding to $-w_{\text{out}}$.
Activates on \textit{Vietnam}

\textbf{4.7667}: all three weights related to consoles in different ways.
Activates on \textit{Xbox}

\textbf{4.9983}: $w_{\text{out}}$ is related to electronic devices, $w_{\text{in}}$ either electronic devices or sports (surfing may belong to both), $w_{\text{gate}}$ is also mostly related to electronic devices.
When using $W_E$, we find $w_{\text{out}}$ contains \textit{iPhone} as a top token.
Activates on \textit{iPhone}.

\textbf{4.10859}: When using $W_E$, we find $w_{\text{gate}}, w_{\text{out}}$ include \textit{Thailand} as a top token, $w_{\text{out}}$ additionally \textit{Buddha, Buddhist}.
Activates on \textit{Thailand}.

\textbf{4.10882}: When using $W_E$, we find $-w_{\text{out}}$ contains \textit{Italy}, $-w_{\text{in}}, w_{\text{gate}}$ additionally contain \textit{Rome}.
Activates on \textit{Italy}.

\textbf{4.10995}: \textit{Boston} appears in gate and \textit{Massachusetts} in $-w_{\text{in}}$.
	When using $W_E$, we find $-w_{\text{out}}, w_{\text{gate}}$ contain \textit{Massachusetts} and \textit{Boston}, $-w_{\text{in}}$ contains \textit{Boston}.
	Activates on \textit{Massachusetts}.

\textbf{22.2589}: $w_{\text{gate}}$ and $-w_{\text{in}}$ recognize tokens like \textit{Islam, Muhammad} and others related to the Arabo-Islamic world. The same goes for $-w_{\text{out}}$ (as it is similar to $w_{\text{in}}$).
Activates on \textit{Muhammad}.

\textbf{24.4880}: For all three weight vectors the first four tokens (but not more) are Philippine-related (even though the gate vector is actually not very similar to the others). The gate vector also reacts to other geographical names, which \textit{may} have in common that they are associated with non-''white'' (Black, Asian or Latin) people in the US sense (\textit{Singapore, Malaysian, Nigerian, Seoul, Pacific, Kerala, Bangkok}, but also \textit{(Los) Angeles} and \textit{Bronx}).
Activates on \textit{Philippines}.
	
\textbf{24.6771}: $w_{\text{gate}}, -w_{\text{in}}, -w_{\text{out}}$ all correspond to capitalized first names.
Activates on \textit{Muhammad}.

\textbf{25.2723}: Some tokens associated with $w_{\text{in}}$ and $w_{\text{out}}$ are possible completions for \textit{th} (\textit{th-ousand, th-ought, th-orn}.
When using $W_E$, in all three weights there are a few \textit{th} tokens, but also with \textit{ph} and similar.
Activates on \textit{Thailand}.

\textbf{25.10496}:
$-w_{\text{in}}, -w_{\text{out}}$ correspond to tokens starting with \textit{v}
(upper or lower case, with or without preceding space).
$w_{\text{gate}}$ on the other hand seems to react to appropriate endings for tokens starting in \textit{v}:
\textit{vol-atility, v-antage, v-intage, vel-ocity, V-ancouver}.
When using $W_E$, we also find all three weight-vectors are very \textit{v}-heavy.
Activates on \textit{Vietnam}.
\enote{sg}{Check:
A possible role for this neuron in our context is the following:
The residual stream contains both \textit{Vietnam} and possible endings for this token
(perhaps \textit{-ese} for Vietnamese).
The neuron therefore activates (negatively) and promotes tokens starting with \textit{v},
which may include the attribute \textit{Vietnamese}.
}
 
\subsection{Depletion neurons}
\textbf{30.9996}: Downgrades weird tokens if present / promotes frequent English stopwords if absent.
Also an attention deactivation neuron for 15 heads in layer 31.


\subsection{Proportional change neurons}

\textbf{25.7032}: Some tokens associated with $w_{\text{gate}}$ and $w_{\text{out}}$ are possible completions for \textit{x} or \textit{ex} (\textit{X-avier, x-yz, ex-cel, ex-ercise}.
When using $W_E$, both \textit{x} and \textit{box} (with variants) appear in all three weight vectors.
Activates on \textit{Xbox}.

\textbf{25.8607}:
All three vectors correspond to tokens related to cities.
Moreover, $-w_{\text{out}}$ seems to correspond to non-city places, such as national governments or villages.
$w_{\text{in}}$ is actually not that similar to $w_{\text{gate}},w_{\text{out}}$ (in terms of cosine similarities),
but all three correspond to city-related tokens. 
When using $W_E$, in all three weights there are a few city-related tokens.
Activates on \textit{Paris}.
We may think of the two input directions as two largely independent ways of checking that ``it's about a city''
(this is a recurring phenomenon that we describe in \cref{sec:doublecheck}).
When the gate activates but the linear input does not confirm it's about a city, the output promotes closely related but non-city interpretations (for example \textit{Paris} actually refers to the French government in some contexts).

\textbf{29.8118}:
\enote{sg}{Check}
Partition neuron, highest variance of all proportional change neurons.
Also an attention deactivation neuron for 4 heads (0,2,11,15) in layer 30.

\textbf{31.5490}:
Activates on \textit{Muhammad}.
$w_{\text{gate}}$ reacts to various Asian names and Asian-sounding subwords,
$w_{\text{in}}$ to surnames as opposed to other English words starting with space and uppercase letter.
\enote{sg}{Check: so the neuron is presumably activated positively.}
$w_{\text{out}}$ corresponds to more Asian stuff (mostly subwords) as opposed to English surnames.

\textbf{31.6275}:
Mostly promotes two-letter tokens (no preceding space, typically uppercase).
$-w_{\text{in}}$ typically lowercase single letters.
$-w_{\text{gate}}$ mostly lowercase two-letter tokens.
``If no lowercase two-letter tokens, promote uppercase two-letter tokens proportionally to absence of lowercase single letters" ?

\textbf{31.8342}: This is an \textit{-ot-} neuron:
$w_{\text{gate}}$ and $w_{\text{out}}$ correspond to \textit{-o(t)-} suffixes,
$-w_{\text{in}}$ to various \textit{-ot-} stuff.
Judging by the weight similarities, we expect that $w_{\text{out}}$ is typically activated negatively:
downgrade \textit{-o(t)-} suffixes if present in the residual stream.
Activates on \textit{Egypt}.

\subsection{Orthogonal output neurons}
\enote{sg}{I should check those}

\textbf{0.1758}: When using $W_E$, all three weight vectors' top tokens are famous web sites, including \textit{YouTube}.
Activates on \textit{YouTube}.

\textbf{0.3338}: When using $W_E$, we find especially $w_{\text{gate}}$ and $-w_{\text{in}}$, but also $-w_{\text{out}}$ are similar to smartphone-related tokens.
Activates on \textit{iPhone}.

\textbf{0.3872}: When using $W_E$, we find especially $w_{\text{gate}}$, but also $-w_{\text{in}}$ and $-w_{\text{out}}$ correspond to city names.
Activates on \textit{Paris}.

\textbf{0.7829}: When using $W_E$, we find $w_{\text{in}}, w_{\text{out}}$ and to a lesser extent $w_{\text{gate}}$ correspond in large part to software names.
Activates on \textit{iTunes}.

\textbf{0.7966}: When using $W_E$, the weight vectors mostly correspond to tokens starting with \textit{th}.
Activates on \textit{Thor}.

\textbf{29.2568}: $w_{\text{out}}$ Asian (Thai?) sounding syllables vs. (Asian) geographic names in English and other stuff; $w_{\text{in}}$ reacts to Thailand and Asian (geography) stuff as opposed to (mostly) US stuff; $w_{\text{gate}}$ pretty much the same.
Activates on \textit{Thailand}.

\textbf{29.3327}: $w_{\text{gate}}$ mostly reacts to city names (\textit{Paris} being the most important one), -$w_{\text{in}}$ countries and cities, especially in continental Europe (\textit{France} and \textit{Paris} on top) as opposed to stuff related to the former British Empire. Relevant is $-w_{\text{out}}$ which corresponds to pieces of geographical names and especially rivers in France (\textit{Se-ine, Rh-one / Rh-ine, Mar-ne, Mos-elle... Norm-andie, Nancy, commun...}). $w_{\text{gate}}$ and -$w_{\text{in}}$ also react to \textit{river(s)}.
Activates on \textit{Paris}.

\textbf{29.4101}: $w_{\text{gate}}$ and $w_{\text{in}}$ react to \textit{YouTube} (top token!), $w_{\text{out}}$ downgrades it (almost bottom token) and promotes \textit{subscrib*, views, channels} etc.
Activates on \textit{YouTube}.

\textbf{29.6417}: Downgrades \textit{recording} and similar. $w_{\text{gate}}$ and $w_{\text{in}}$ are also similar and involve \textit{iTunes}.
Activates on \textit{iTunes}.

\textbf{29.9734}: $w_{\text{gate}}$ reacts to the East in a broad sense as opposed to the West (\textit{Iran, Kaz-akhstan, Kash-mir, Ukraine}...), $w_{\text{in}}$ mostly to male first names without preceding space. $w_{\text{out}}$ seems to produce word pieces that could begin a foreign name.
Activates on \textit{Muhammad}.
	
\textbf{30.2667}: $w_{\text{gate}}$ reacts to suffixes (for adjectives derived from place names) like \textit{en, ian, ians}, basically the same for $w_{\text{in}}$ and $w_{\text{out}}$.
Activates on \textit{Muhammad}.

\textbf{30.3143}: $w_{\text{gate}}$ reacts to words related to entities that are authoritative for various reasons (\textit{officials, authorities, according, researchers, spokesman, investigators}...). $-w_{\text{in}}$ reacts to uncertainty (\textit{reportedly, according... allegedly... accused}). $-w_{\text{out}}$ is again \textit{police, authorities, officials, court} but with no preceding space.
Activates on \textit{Philippines}.
What authorities and uncertainty have to do with the Philippines is unclear.

\textbf{30.3883}: $w_{\text{gate}}$ and $-w_{\text{in}}$ react to \textit{Virginia} and \textit{Afghanistan}, among others (in the case of $w_{\text{gate}}$: as opposed to other geographical names with no preceding space associated with the South and the sea); $-w_{\text{out}}$ is activated and promotes all variants of \textit{af} (and \textit{ghan}) but downgrades Virginia etc.
Activates on \textit{Afghanistan}.

\textbf{30.4577}: Seems to be related to rugby: $w_{\text{gate}}$ and slightly less obviously $w_{\text{in}}$ react to rugby-related tokens (\textit{midfielder, quarterback}...); $w_{\text{out}}$ promotes different tokens that upon reflection could be related to rugby as well.
Activates on \textit{Ireland}.

\textbf{30.5372}: Promotes \textit{natural} and related, downgrades \textit{inst} tokens. $w_{\text{in}}$ reacts to \textit{wildlife} etc. as opposed to \textit{institute} etc, $w_{\text{gate}}$ reacts to \textit{institute} as opposed to \textit{natural}.
Activates on \textit{Massachusetts}
(in which situation it promotes \textit{Institute}, which makes sense because of MIT).

\textbf{30.8535}: $-w_{\text{out}}$ is \textit{one} in all variants, $w_{\text{gate}}$ too, $w_{\text{in}}$ splits \textit{one, ones} and the equivalent Chinese characters, on the positive side, from \textit{One, 1, ONE} on the negative side (and many other things on both sides).
Activates on \textit{Xbox}.
Presumably this happens because \textit{One} is a possible prediction (\textit{Xbox One}),
and presumably the output reinforces that.

\textbf{31.2135}: orthogonal output, on the conditional enrichment side (weak conditional enrichment, one of the neurons on the vertical axis). $w_{\text{gate}}$ reacts to single letters or symbols as opposed to some English content words without preceding space; $w_{\text{in}}$ and $w_{\text{out}}$ mostly Chinese or Japanese characters as opposed to some Latin diacritics and other weird stuff. Language choice? ``If it's not English and single letters are floating around, make sure to choose the right language / character set."
\enote{sg}{This one may be interesting for the main paper!}

\textbf{31.10424}: $w_{\text{gate}}, -w_{\text{in}}, w_{\text{out}}$ correspond to \textit{score} in the top tokens, which is downgraded if present.
Activates on \textit{Paris}.
No idea what's happening here.

\begin{figure*}
	\subfigure{\includegraphics[width=3.25in]{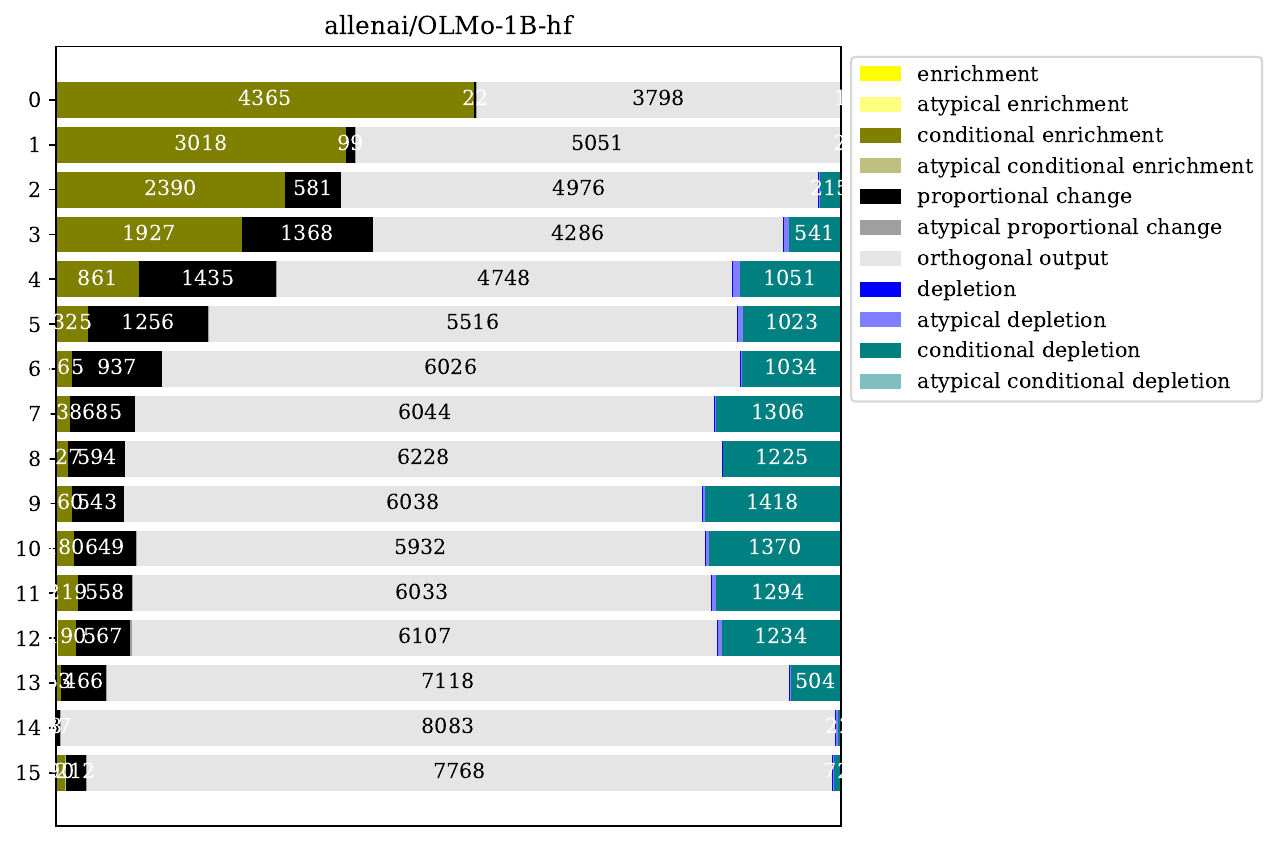}}
	\subfigure{\includegraphics[width=3.25in]{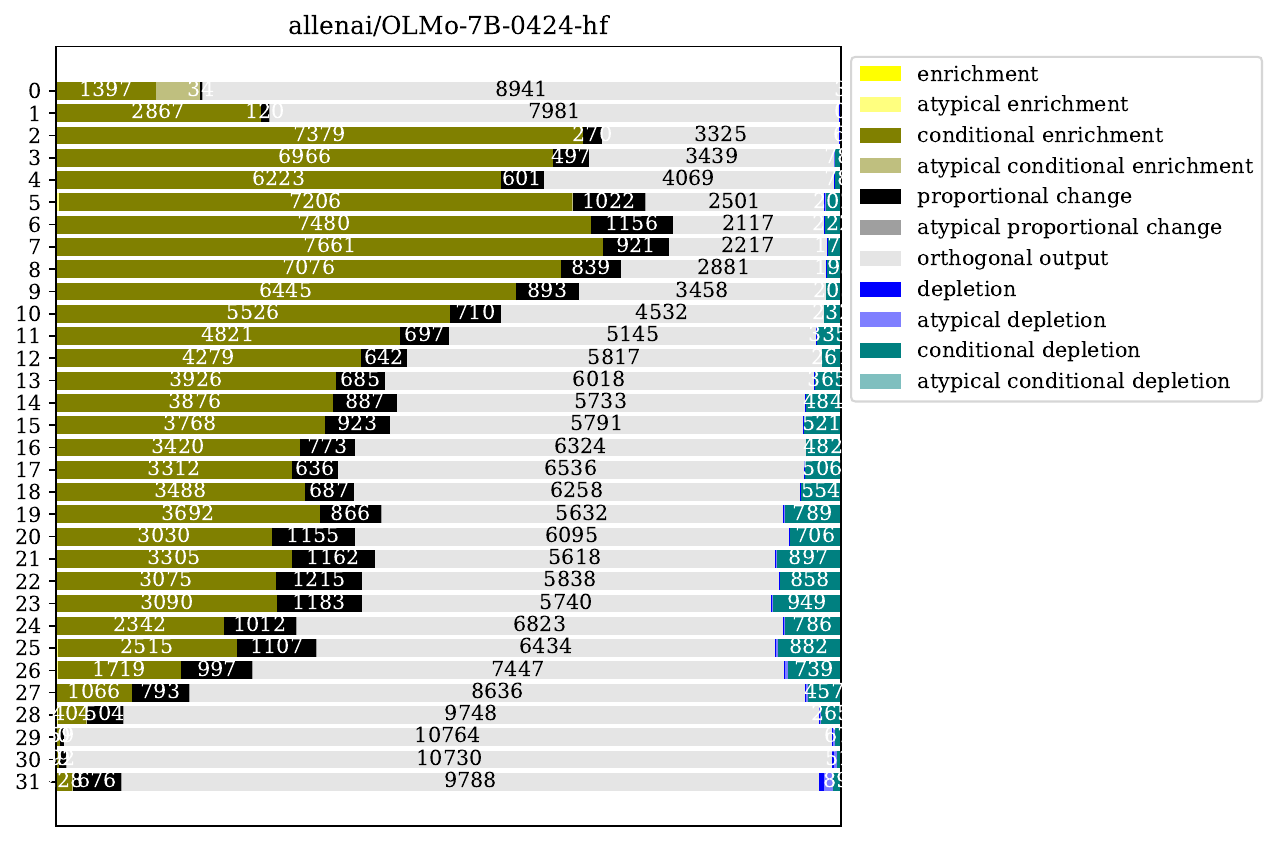}}
	\subfigure{\includegraphics[width=3.25in]{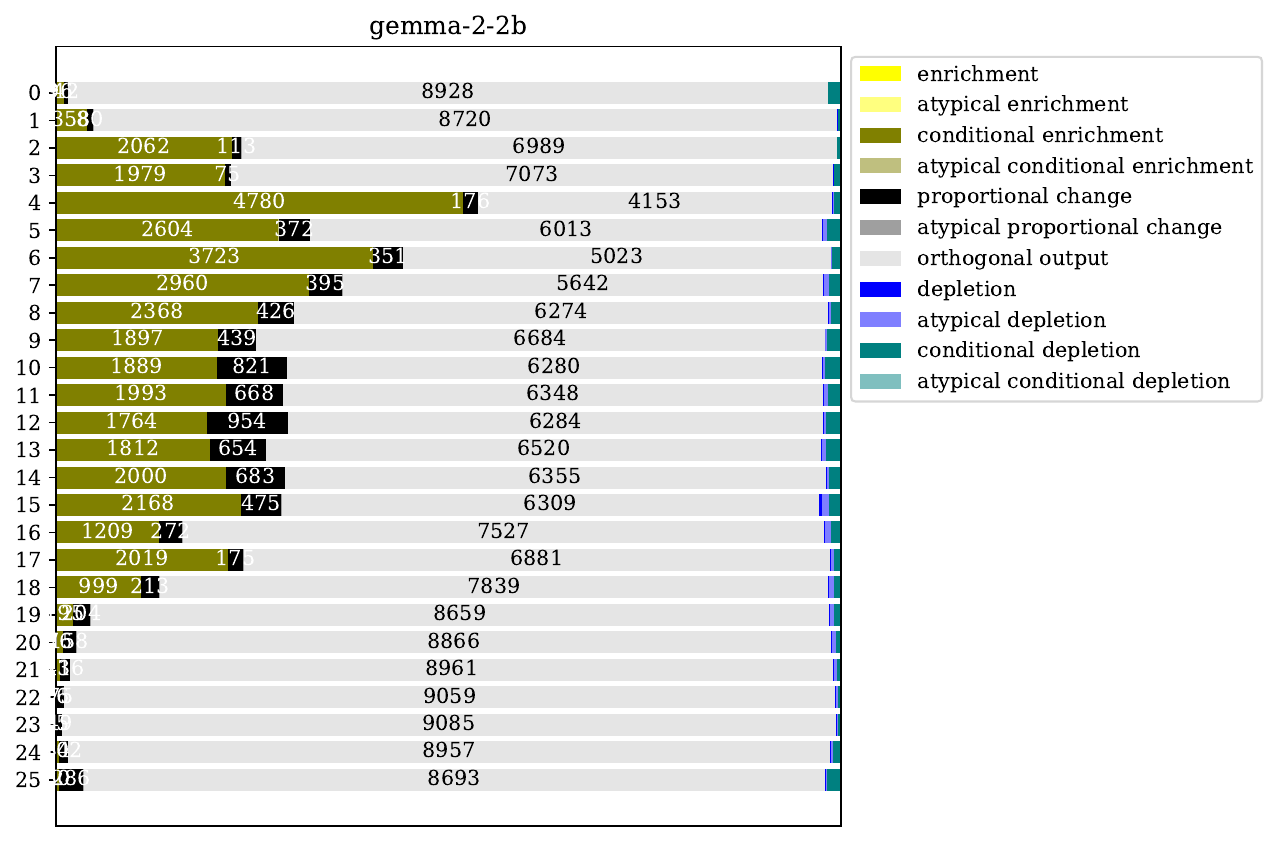}}
	\subfigure{\includegraphics[width=3.25in]{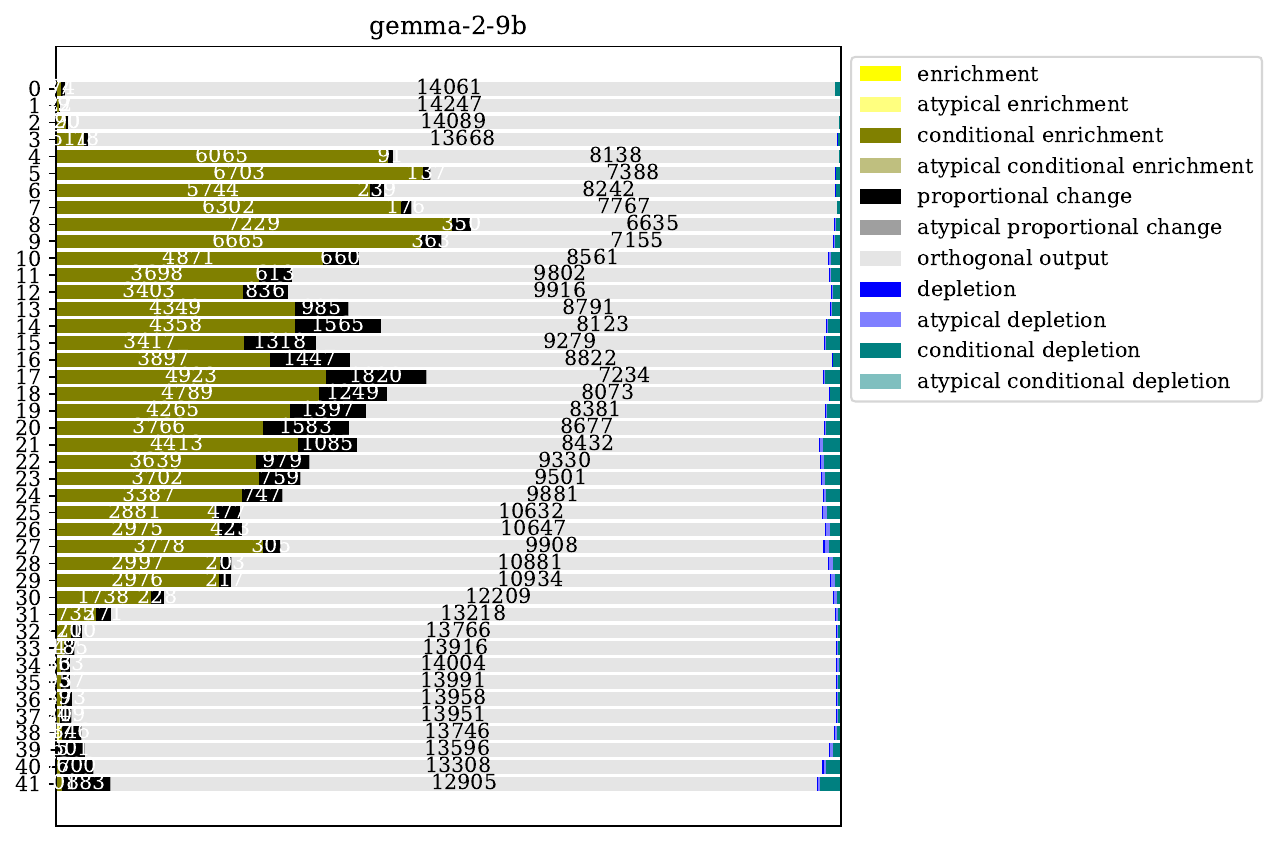}}
	\subfigure{\includegraphics[width=3.25in]{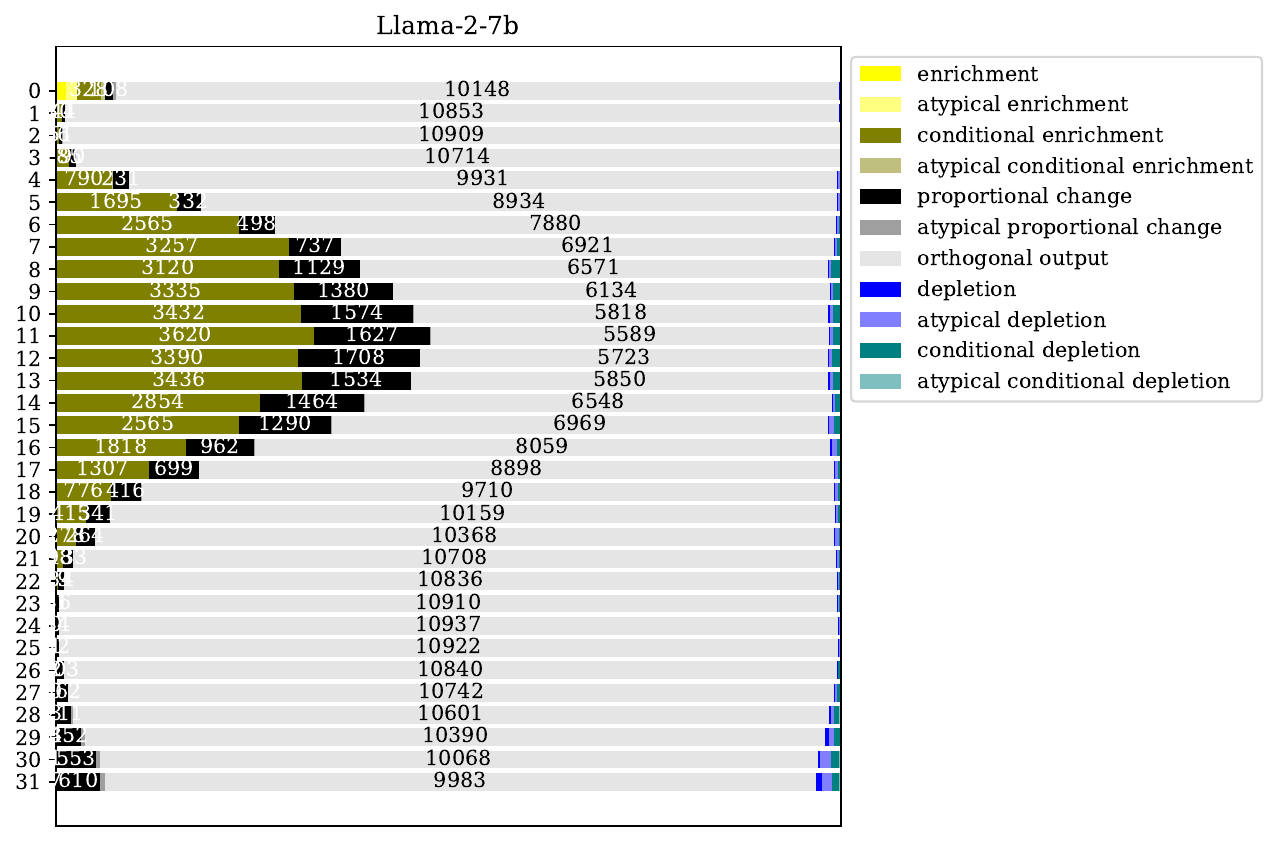}}
	\subfigure{\includegraphics[width=3.25in]{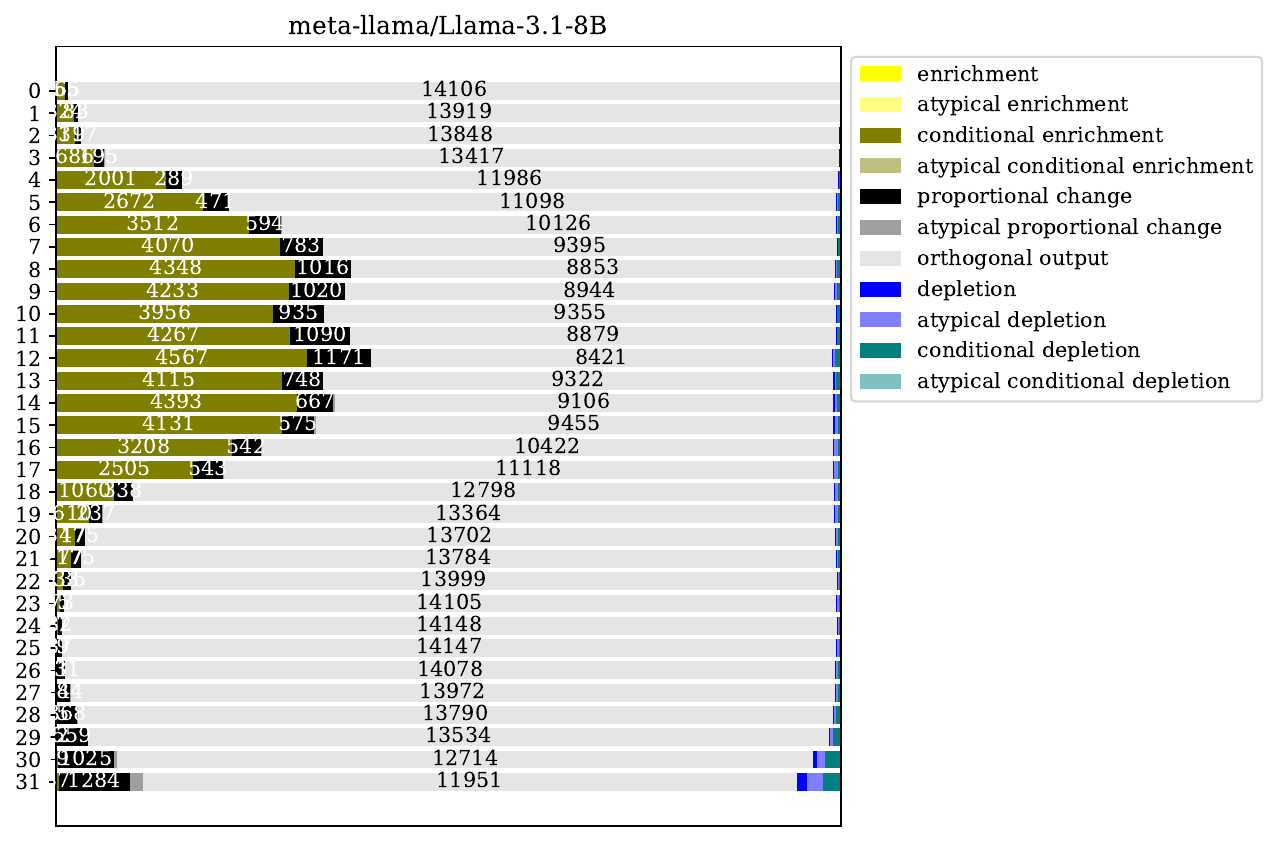}}
	\vskip -0.2in
	\caption{Distribution of neurons by layer and category for a range of models}
	\label{fig:coarse}
\end{figure*}

\begin{figure*}
	\subfigure{\includegraphics[width=3.25in]{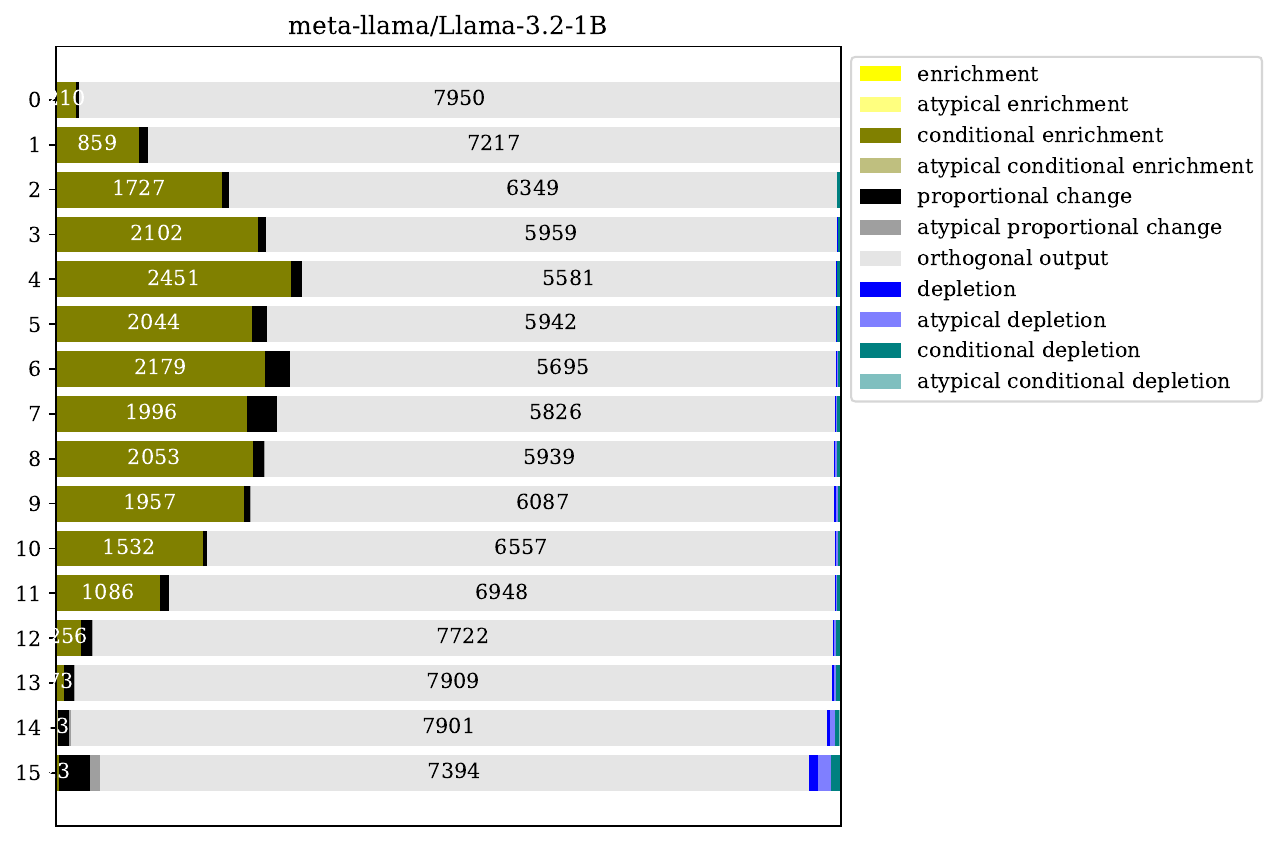}}
	\subfigure{\includegraphics[width=3.25in]{plots/meta-llama/Llama-3.2-3B/coarse.pdf}}
	\subfigure{\includegraphics[width=3.25in]{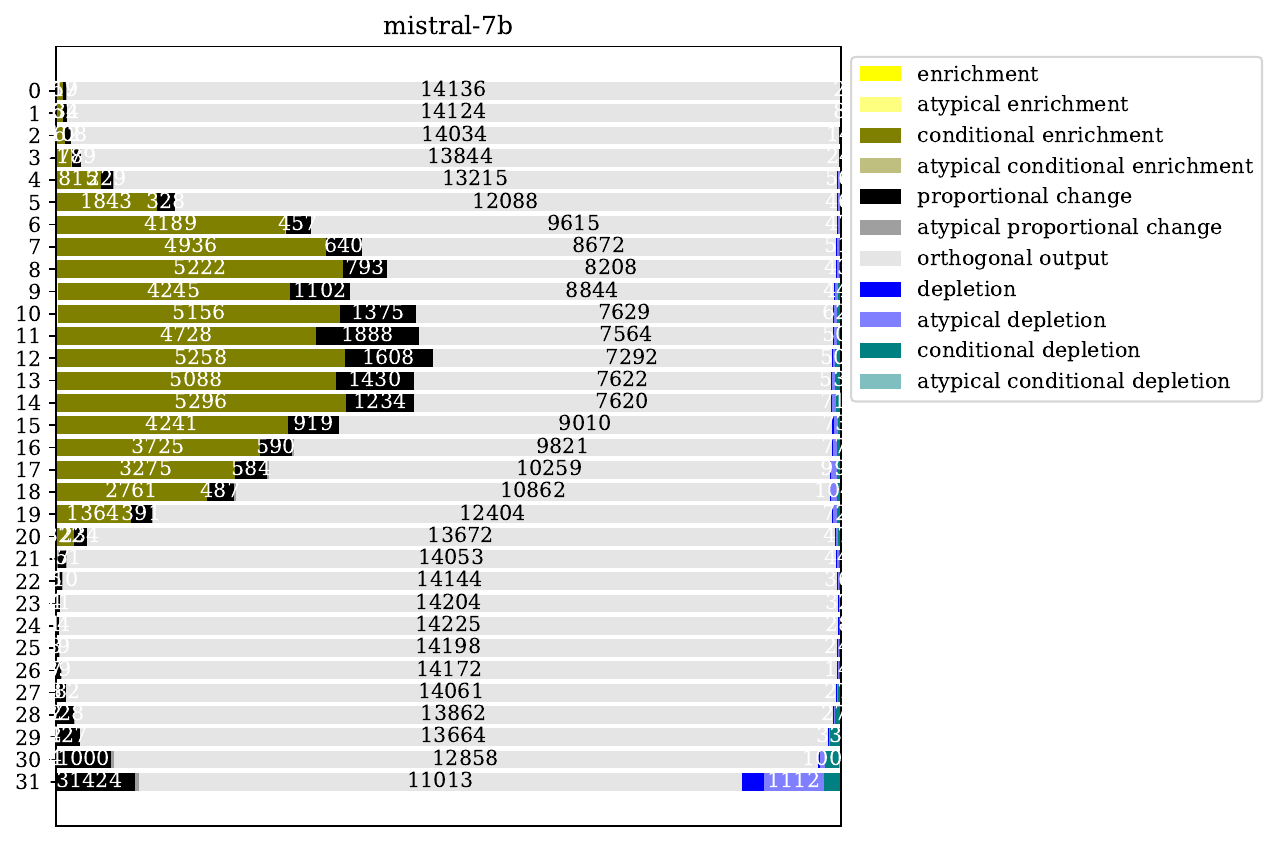}}
	\subfigure{\includegraphics[width=3.25in]{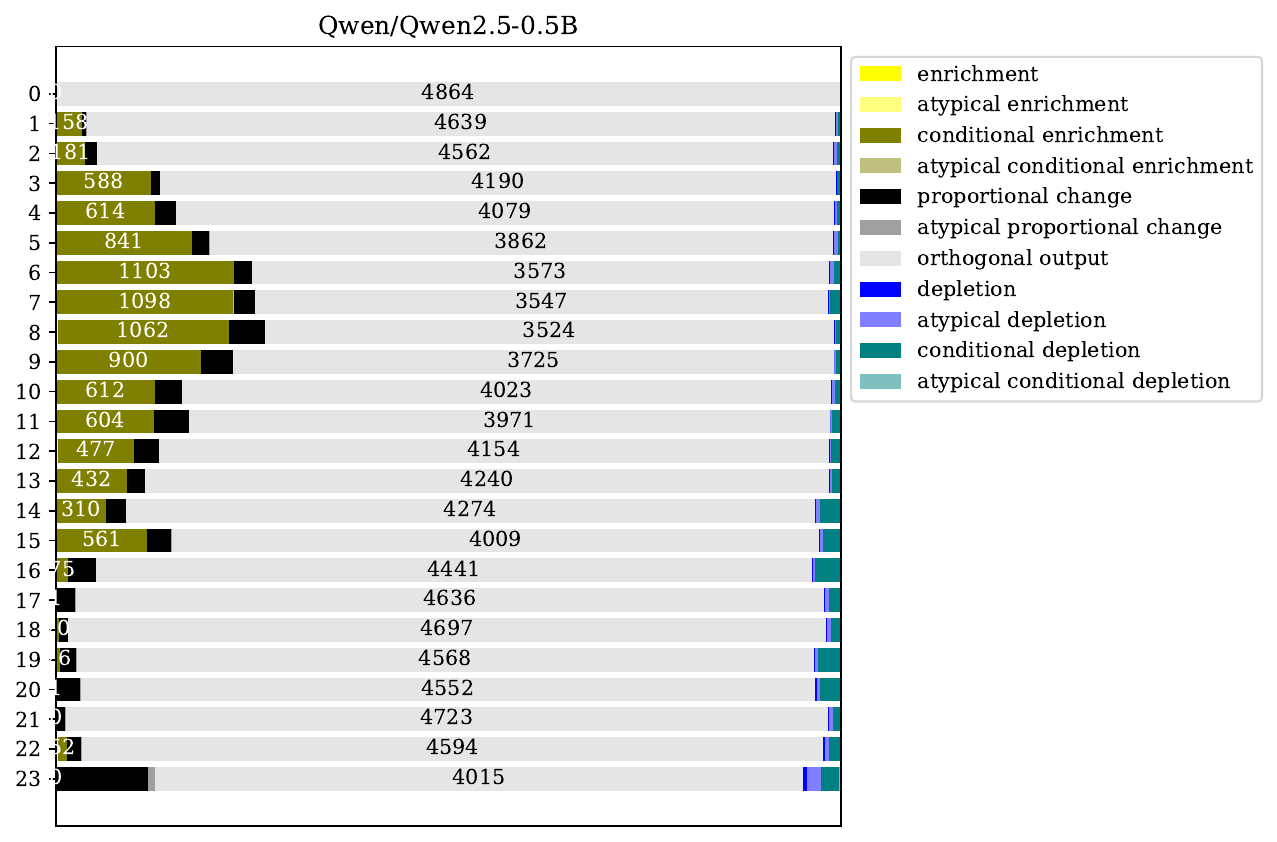}}
	\subfigure{\includegraphics[width=3.25in]{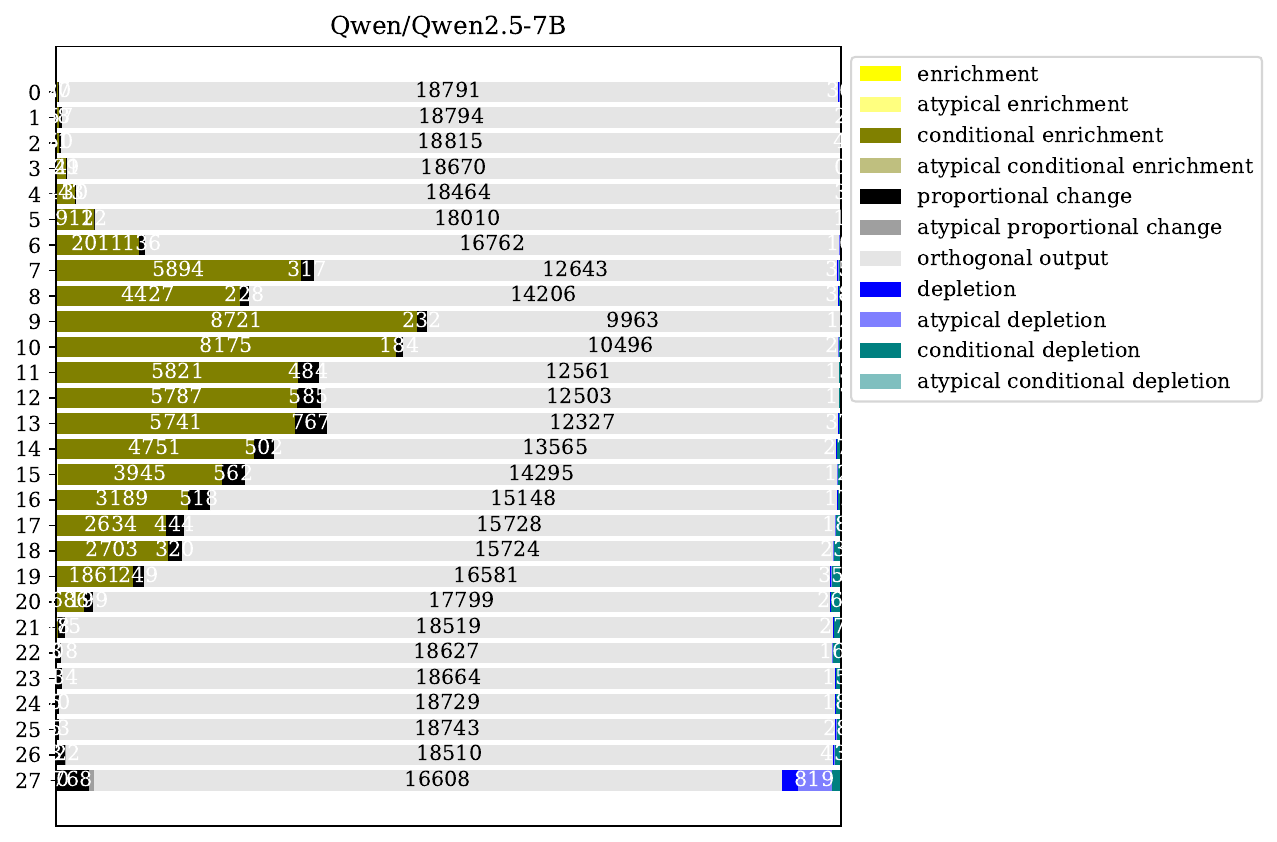}}
	\subfigure{\includegraphics[width=3.25in]{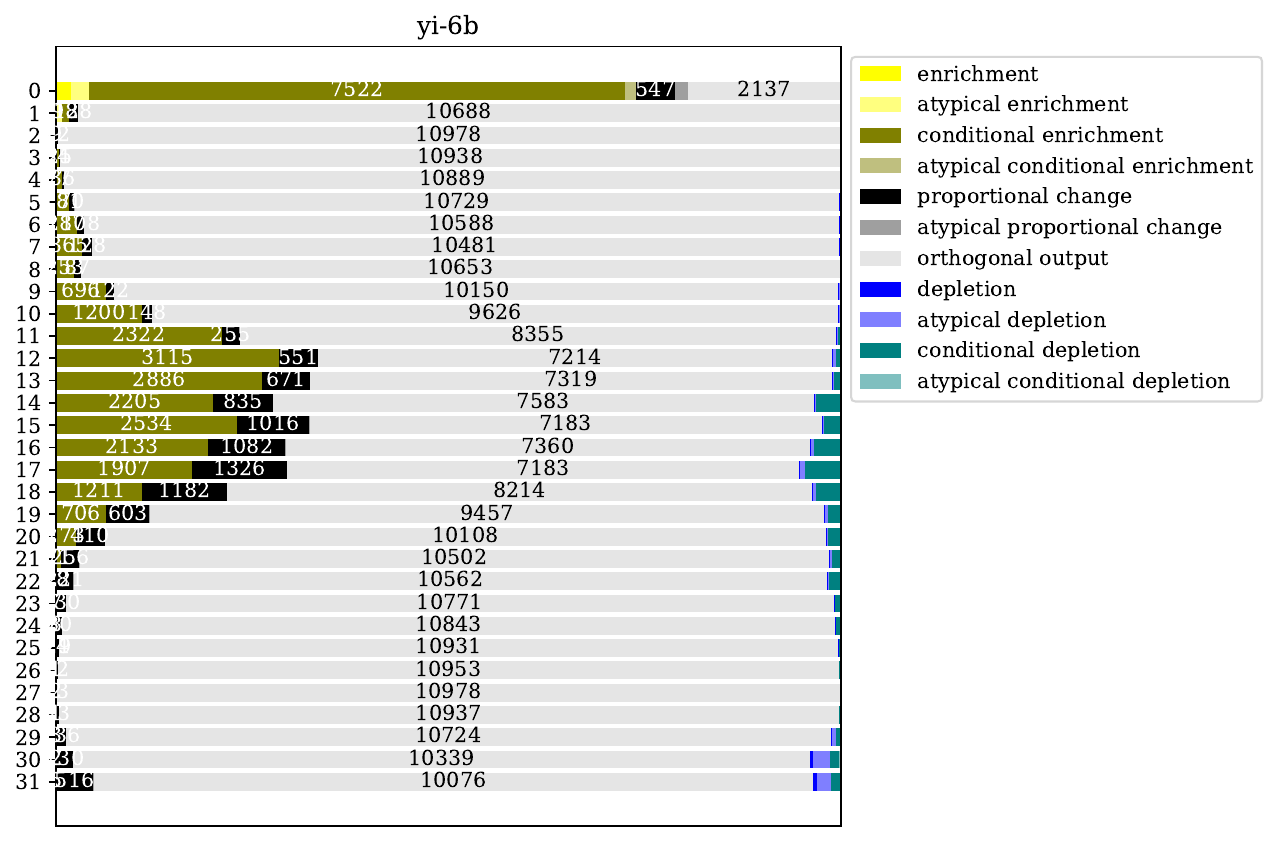}}
	\vskip -0.2in
	\caption{Continuation of \cref{fig:coarse}. Including a copy of \cref{fig:bar} (Llama-3.2-3B) for convenience.}
\end{figure*}

\begin{figure*}
	\subfigure{\includegraphics[width=3.25in]{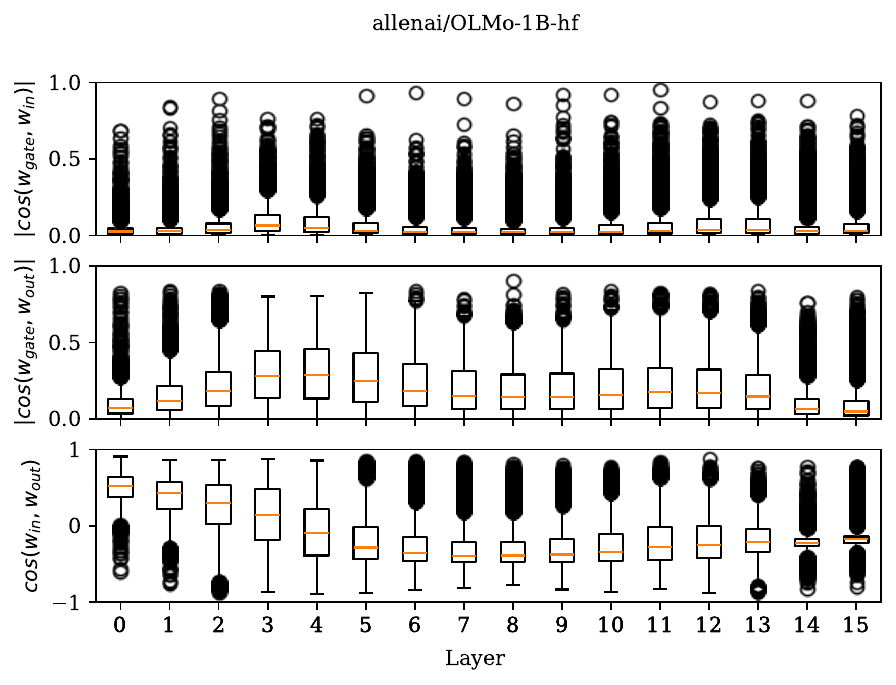}}
	\subfigure{\includegraphics[width=3.25in]{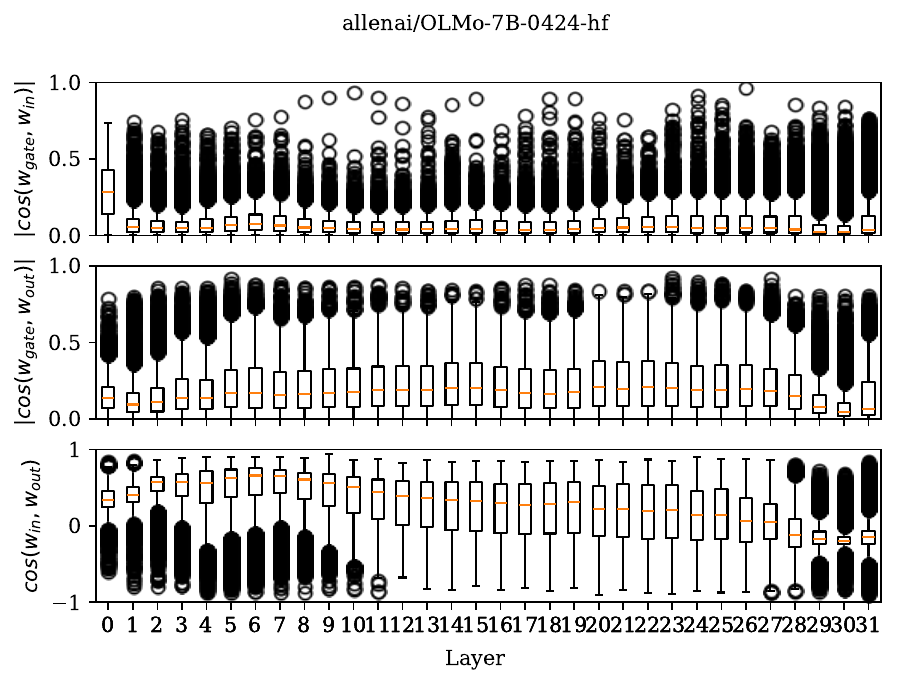}}
	\subfigure{\includegraphics[width=3.25in]{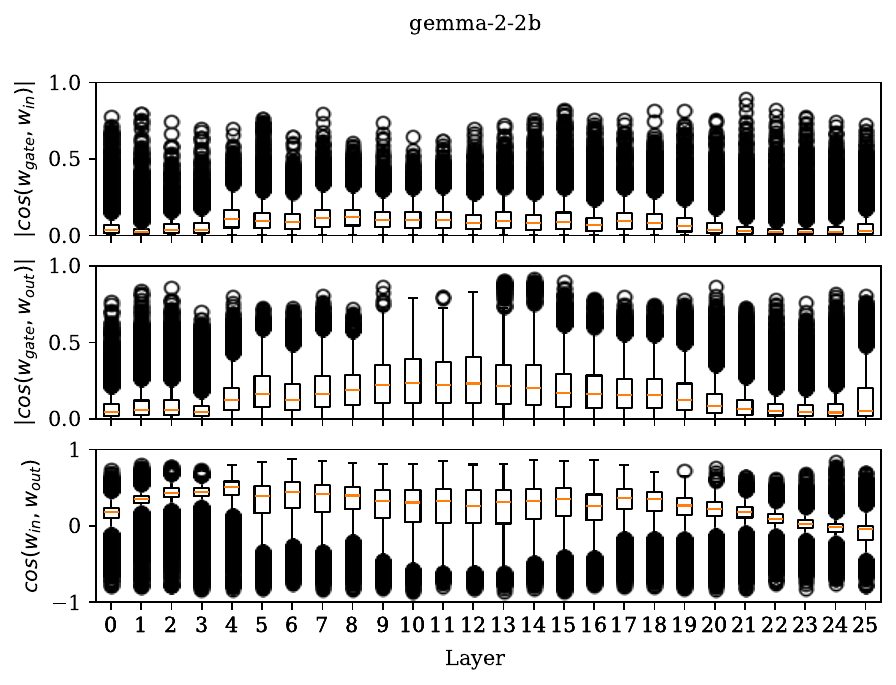}}
	\subfigure{\includegraphics[width=3.25in]{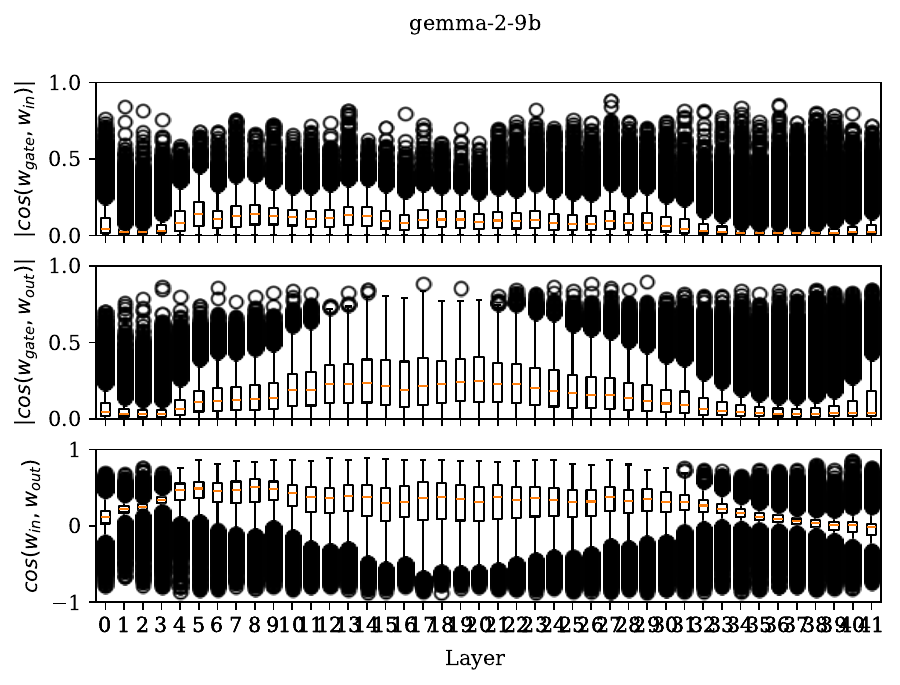}}
	\subfigure{\includegraphics[width=3.25in]{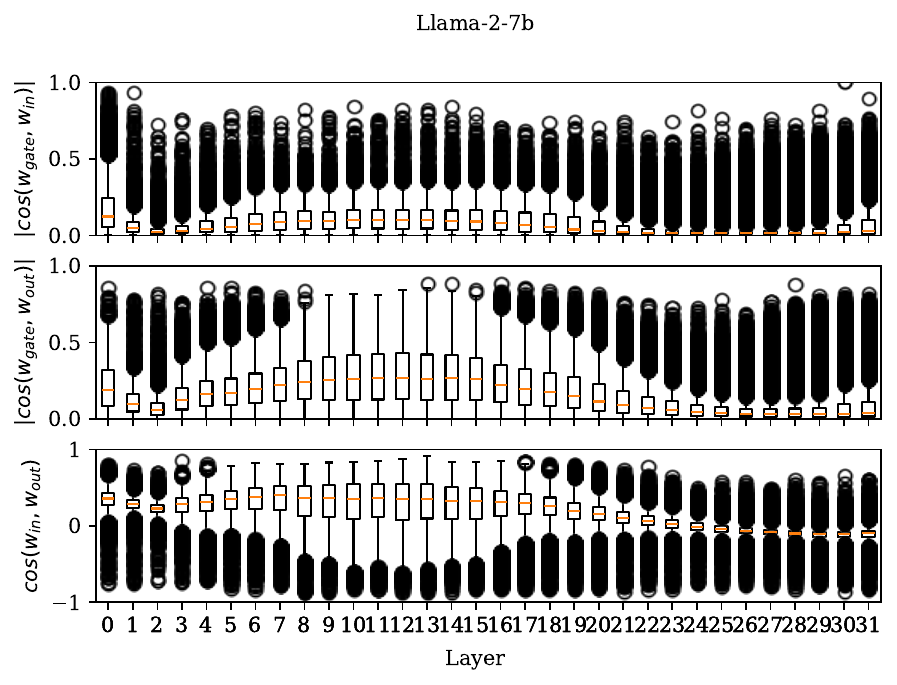}}
	\subfigure{\includegraphics[width=3.25in]{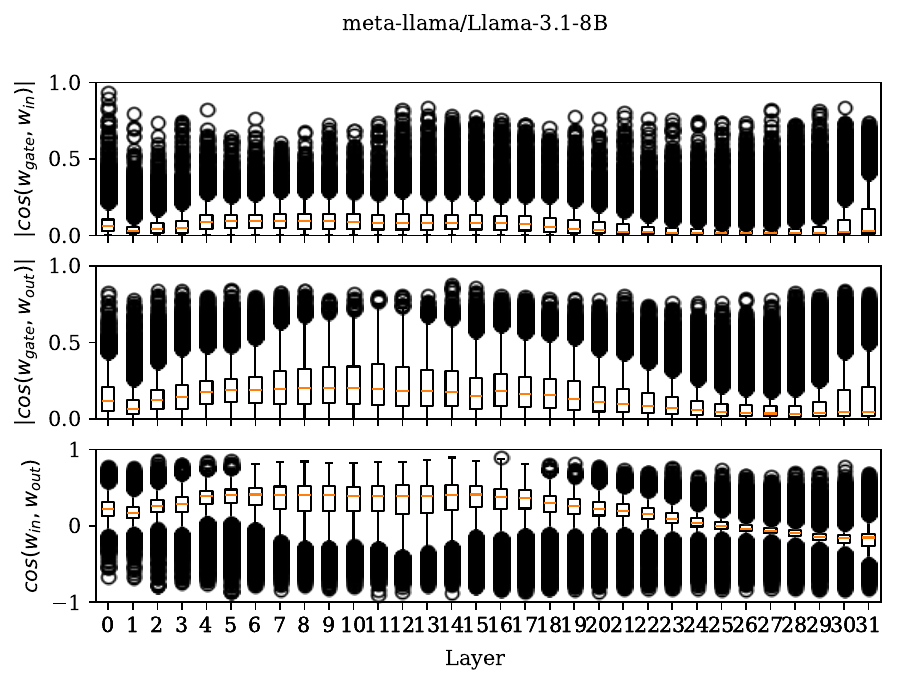}}
	\vskip -0.2in
	\caption{Boxplots for the distribution of weight cosine similarities in each layer. For $\cos(w_{\text{gate}},w_{\text{in}})$ and $\cos(w_{\text{gate}},w_{\text{out}})$ we show the absolute value since their sign does not carry any information on its own. }
	\label{fig:boxplot}
\end{figure*}

\begin{figure*}
	\subfigure{\includegraphics[width=3.25in]{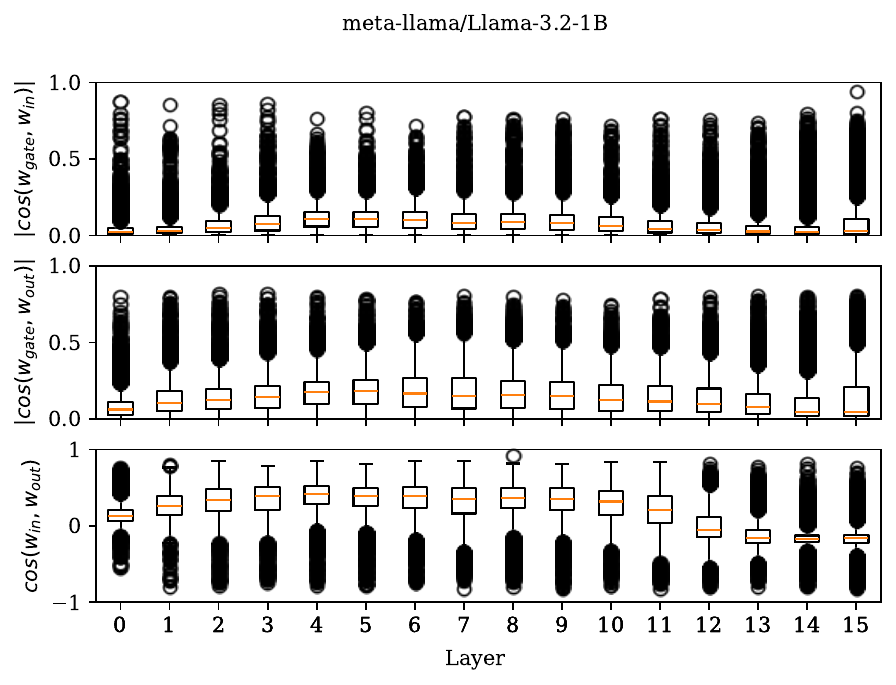}}
	\subfigure{\includegraphics[width=3.25in]{plots/meta-llama/Llama-3.2-3B/boxplot.pdf}}
	\subfigure{\includegraphics[width=3.25in]{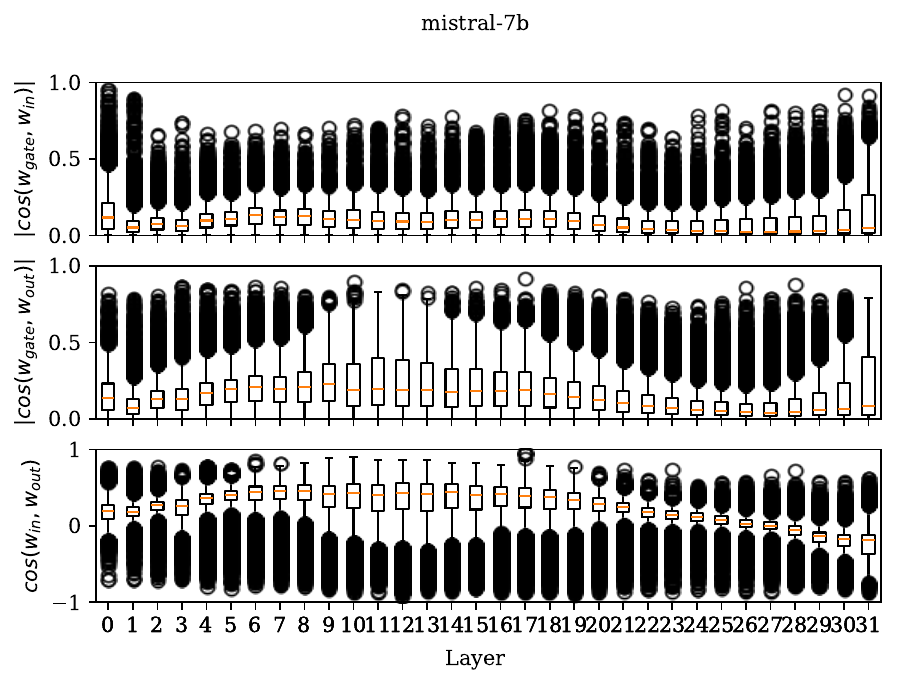}}
	\subfigure{\includegraphics[width=3.25in]{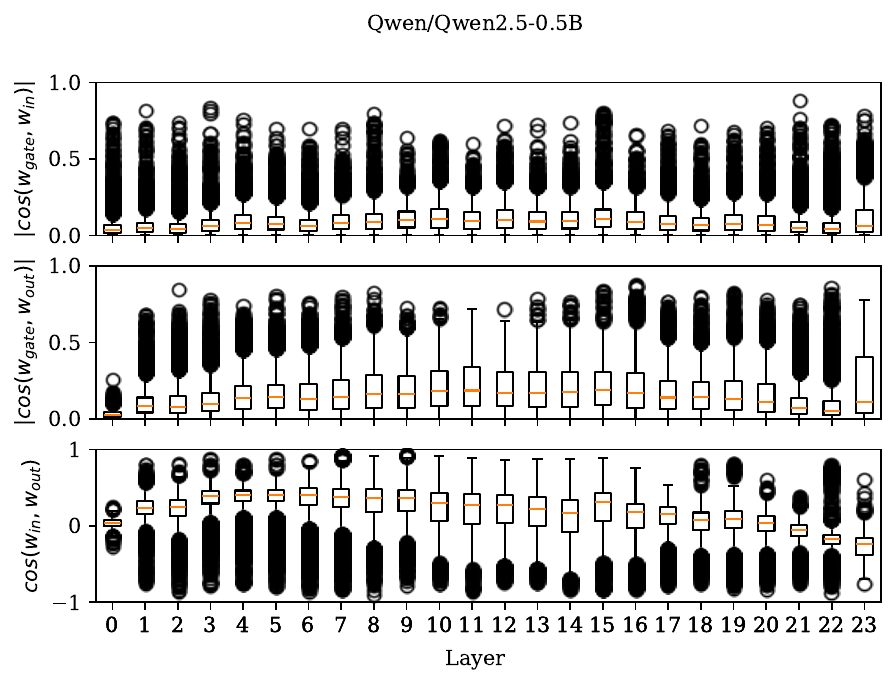}}
	\subfigure{\includegraphics[width=3.25in]{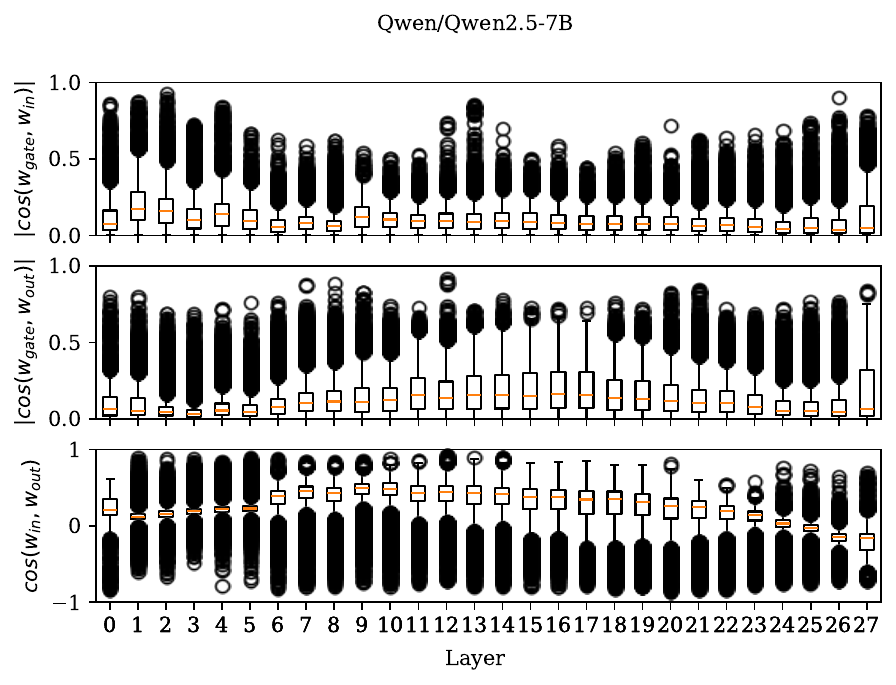}}
	\subfigure{\includegraphics[width=3.25in]{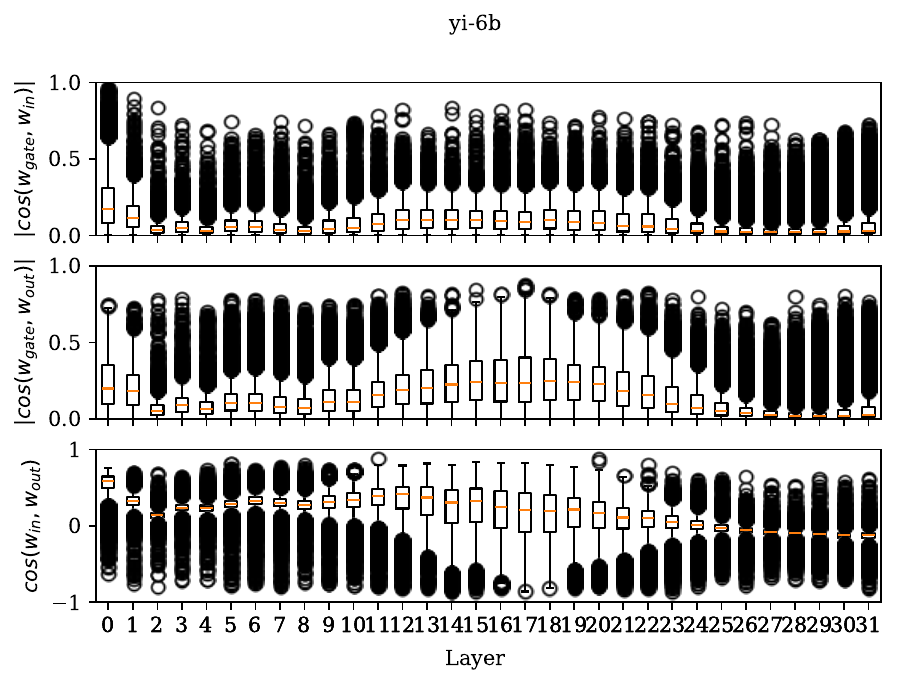}}
	\vskip -0.2in
	\caption{Continuation of \cref{fig:boxplot}. Including a copy of \cref{fig:boxplots} (Llama-3.2-3B) for convenience.}
\end{figure*}

\begin{figure*}
	\centering
	\includegraphics[width=.67\textwidth]{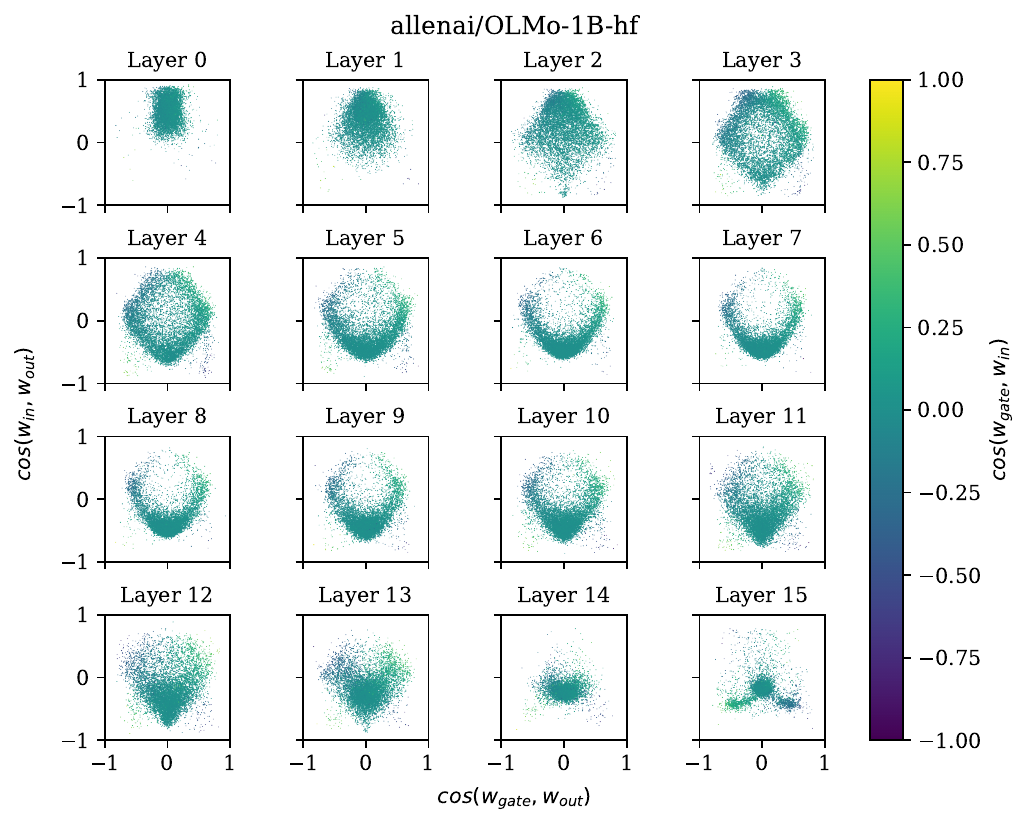}
	\vskip -0.2in
	\caption{}
\end{figure*}
\begin{figure*}
	\centering
	\includegraphics[width=.67\textwidth]{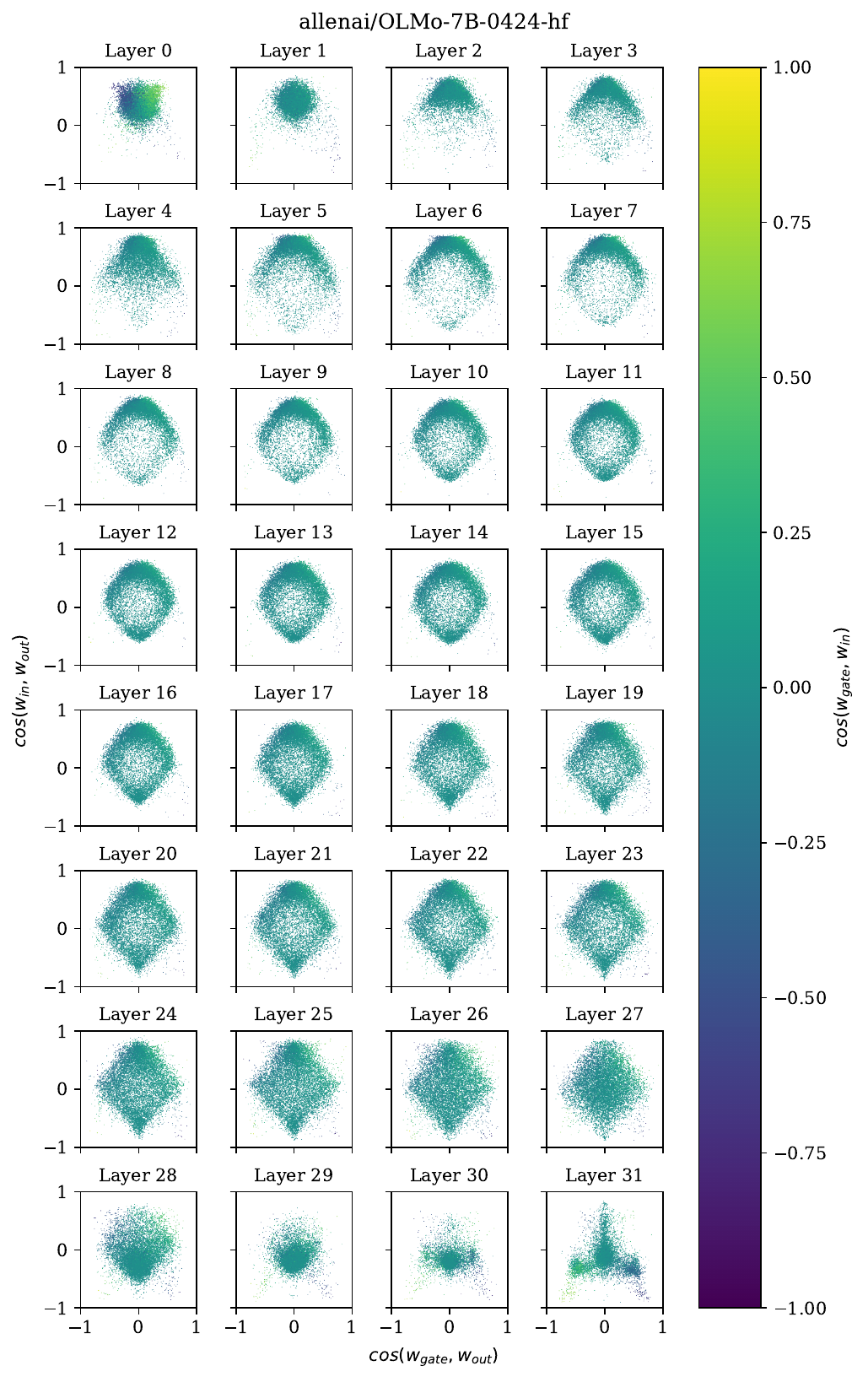}
	\vskip -0.2in
	\caption{}
\end{figure*}
\begin{figure*}
	\centering
	\includegraphics[width=.67\textwidth]{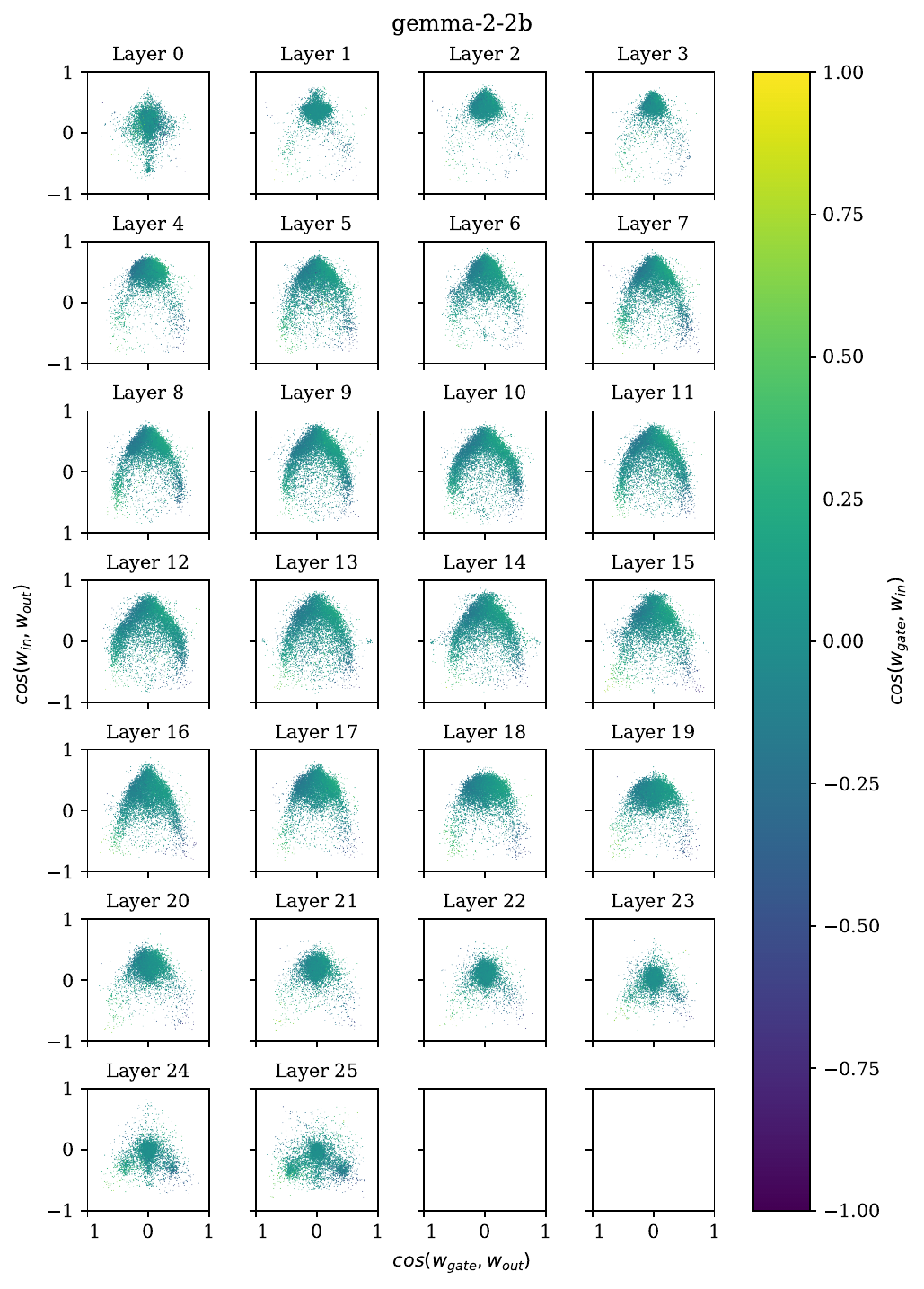}
	\vskip -0.2in
	\caption{}
\end{figure*}
\begin{figure*}
	\centering
	\includegraphics[width=.67\textwidth]{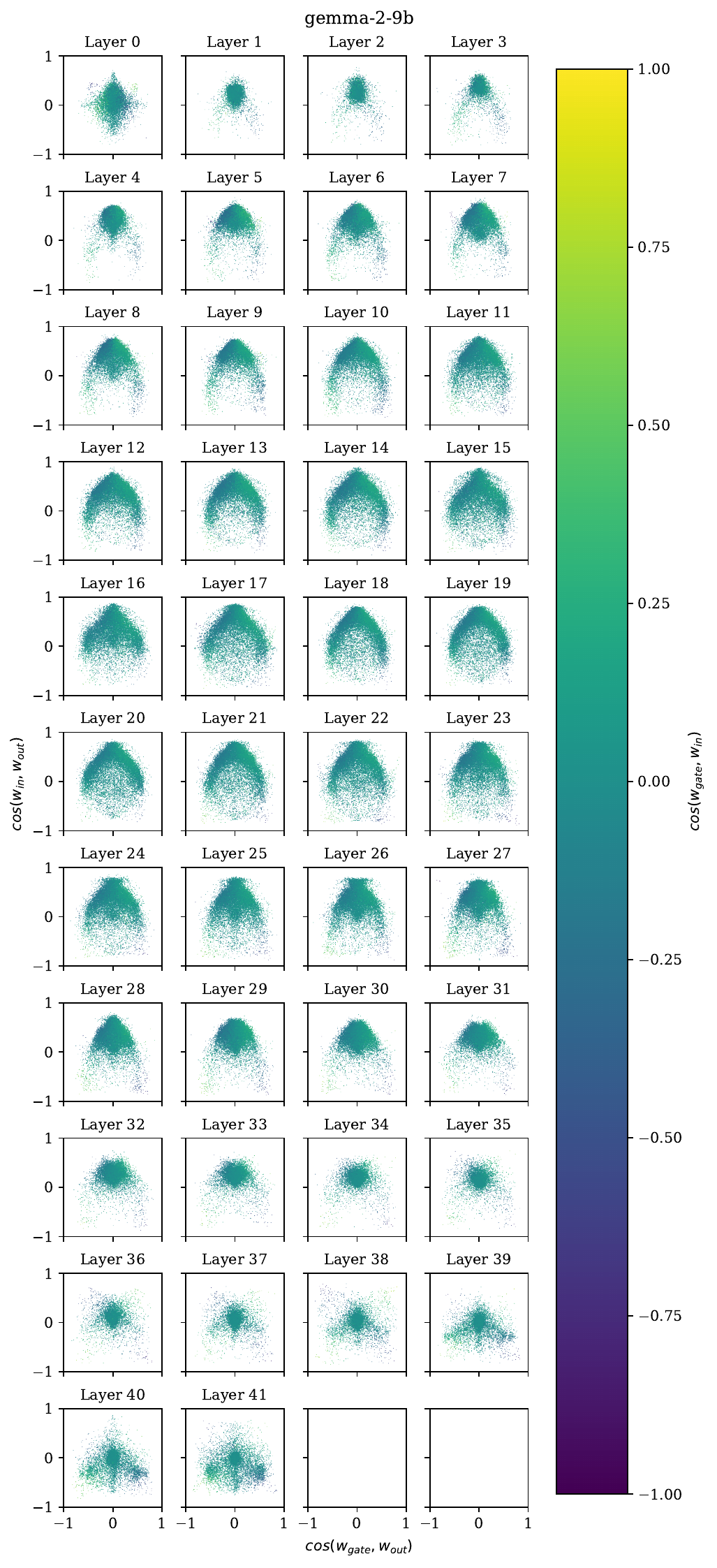}
	\vskip -0.2in
	\caption{}
\end{figure*}
\begin{figure*}
	\centering
	\includegraphics[width=.67\textwidth]{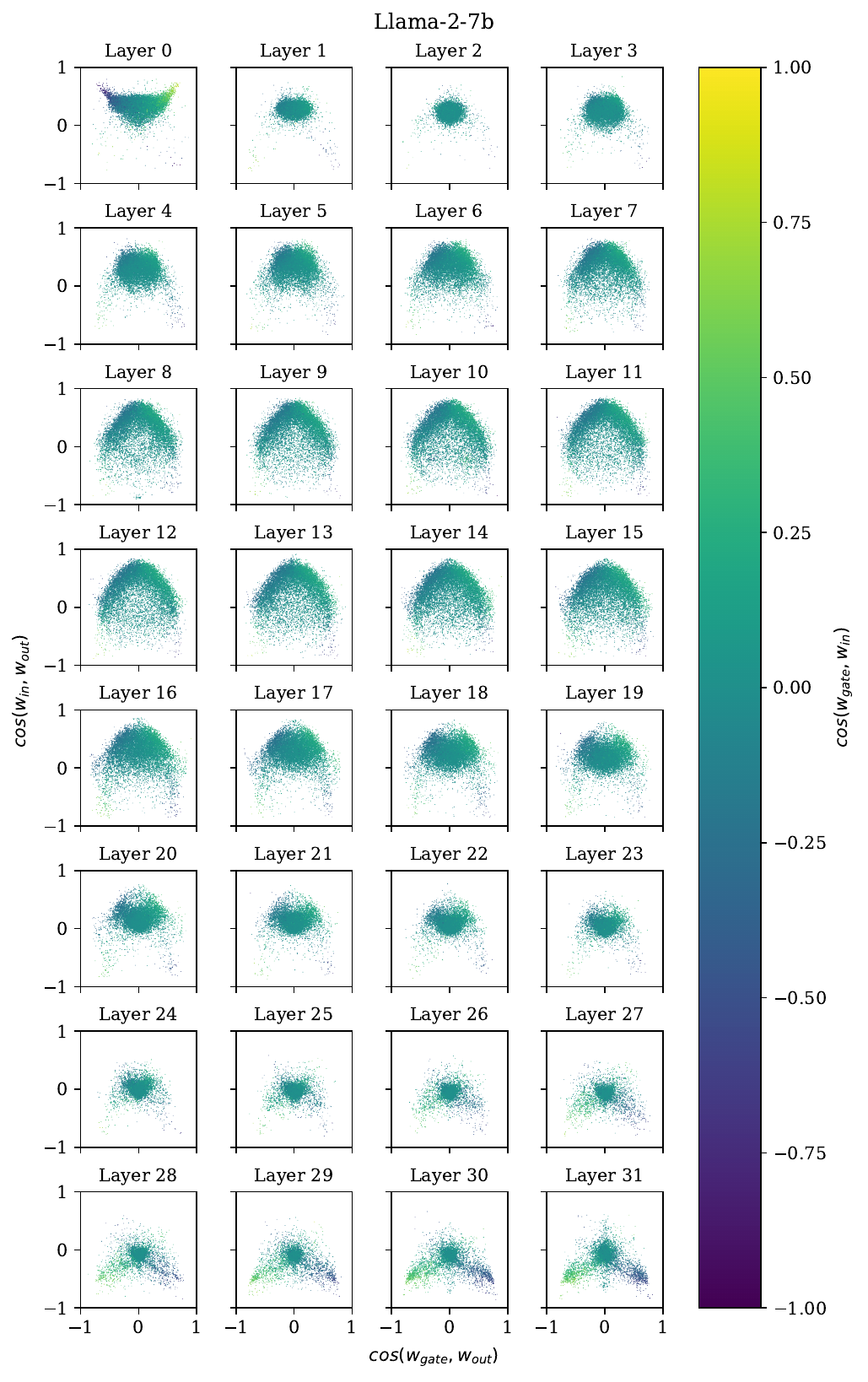}
	\vskip -0.2in
	\caption{}
\end{figure*}
\begin{figure*}
	\centering
	\includegraphics[width=.67\textwidth]{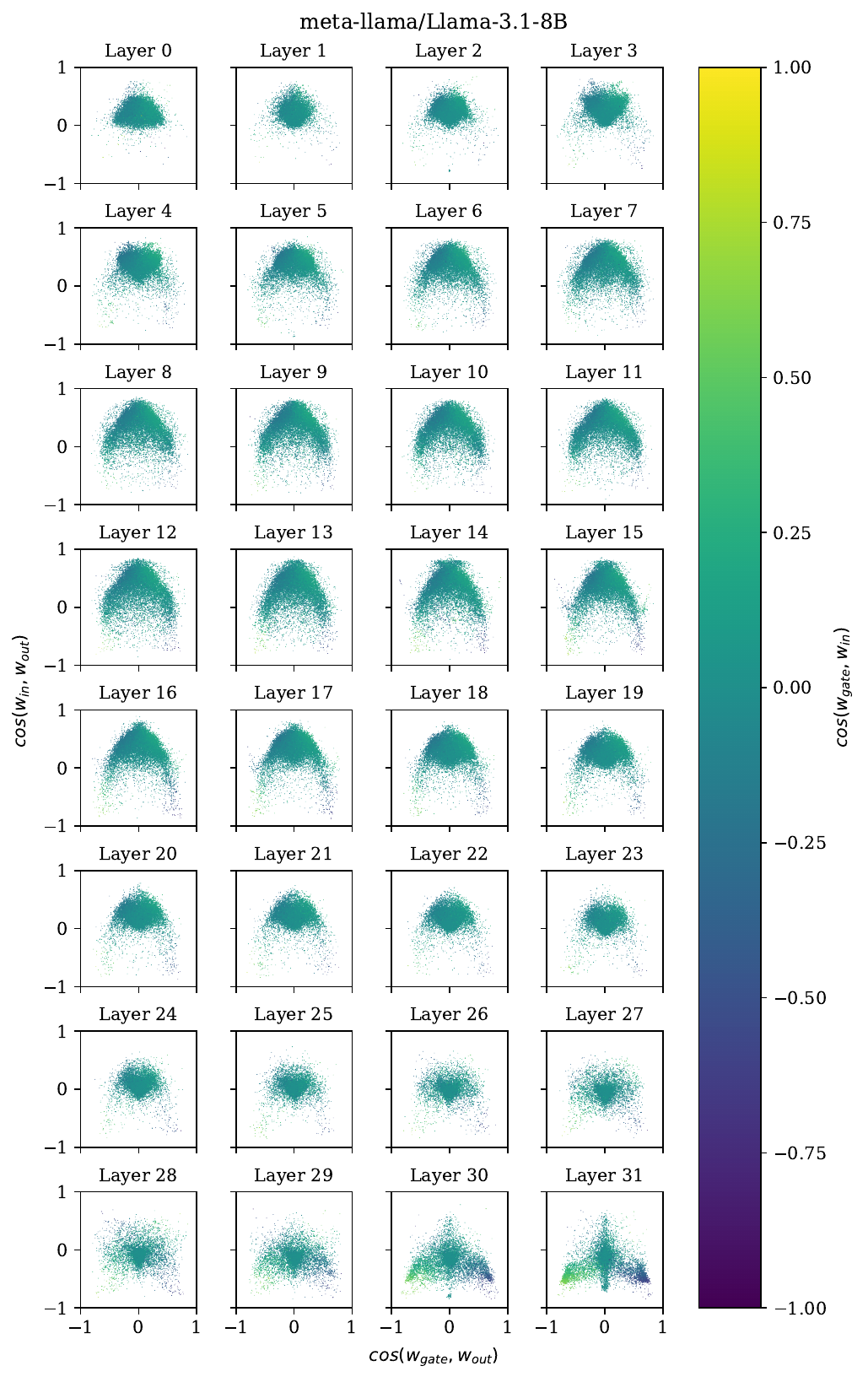}
	\vskip -0.2in
	\caption{}
\end{figure*}
\begin{figure*}
	\centering
	\includegraphics[width=.67\textwidth]{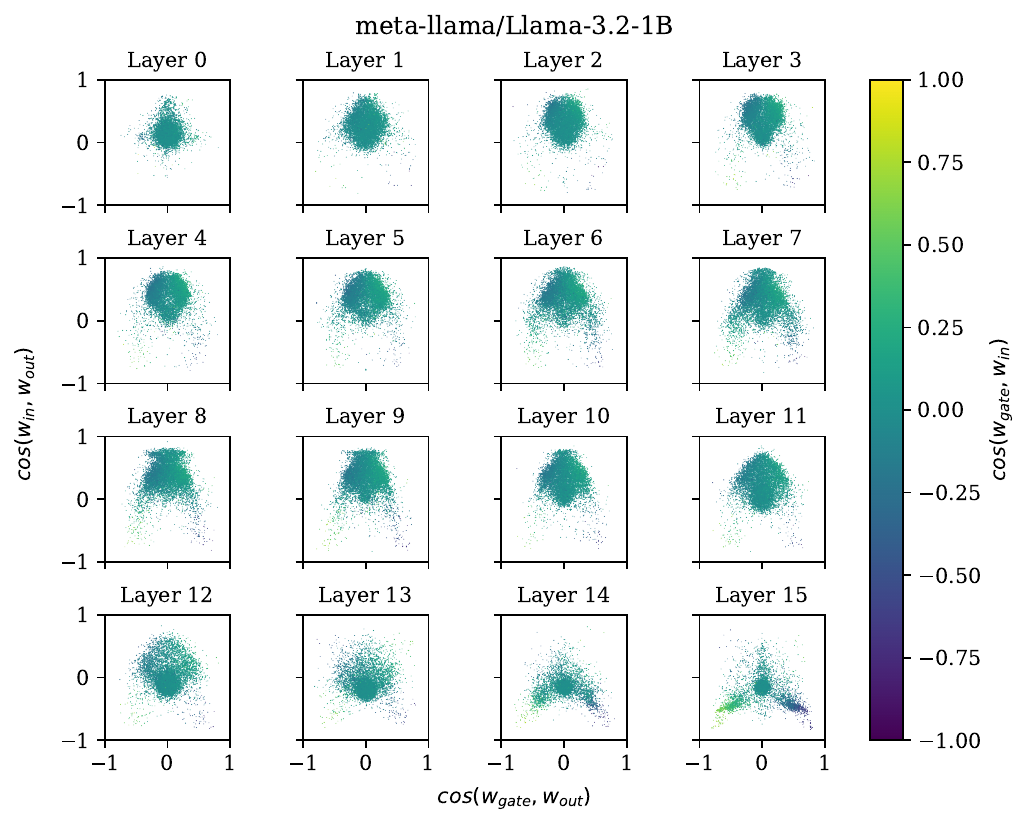}
	\vskip -0.2in
	\caption{}
\end{figure*}
\begin{figure*}
	\centering
	\includegraphics[width=.67\textwidth]{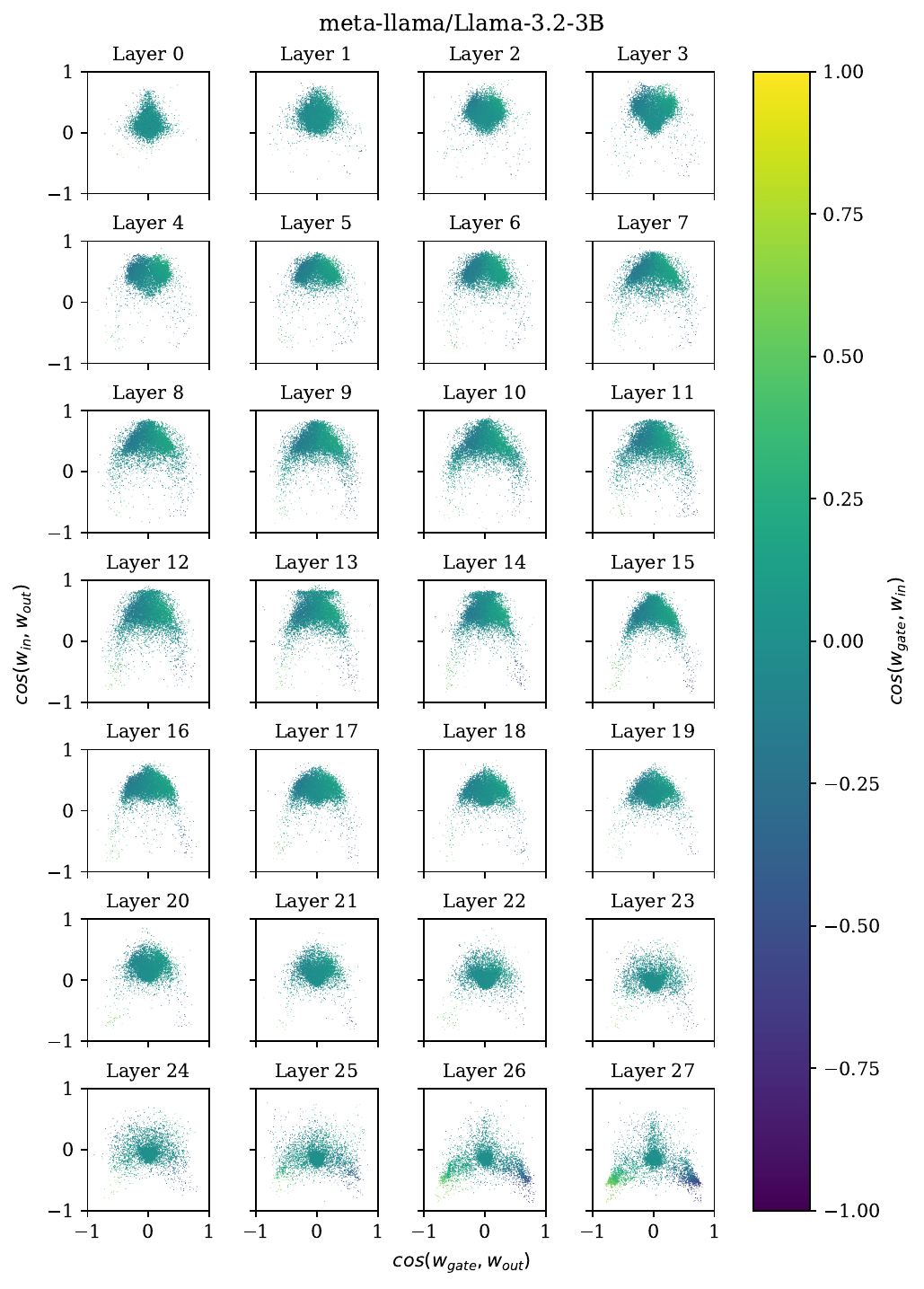}
	\vskip -0.2in
	\caption{Llama-3.2-3B}
	\label{fig:wcos}
\end{figure*}
\begin{figure*}
	\centering
	\includegraphics[width=.67\textwidth]{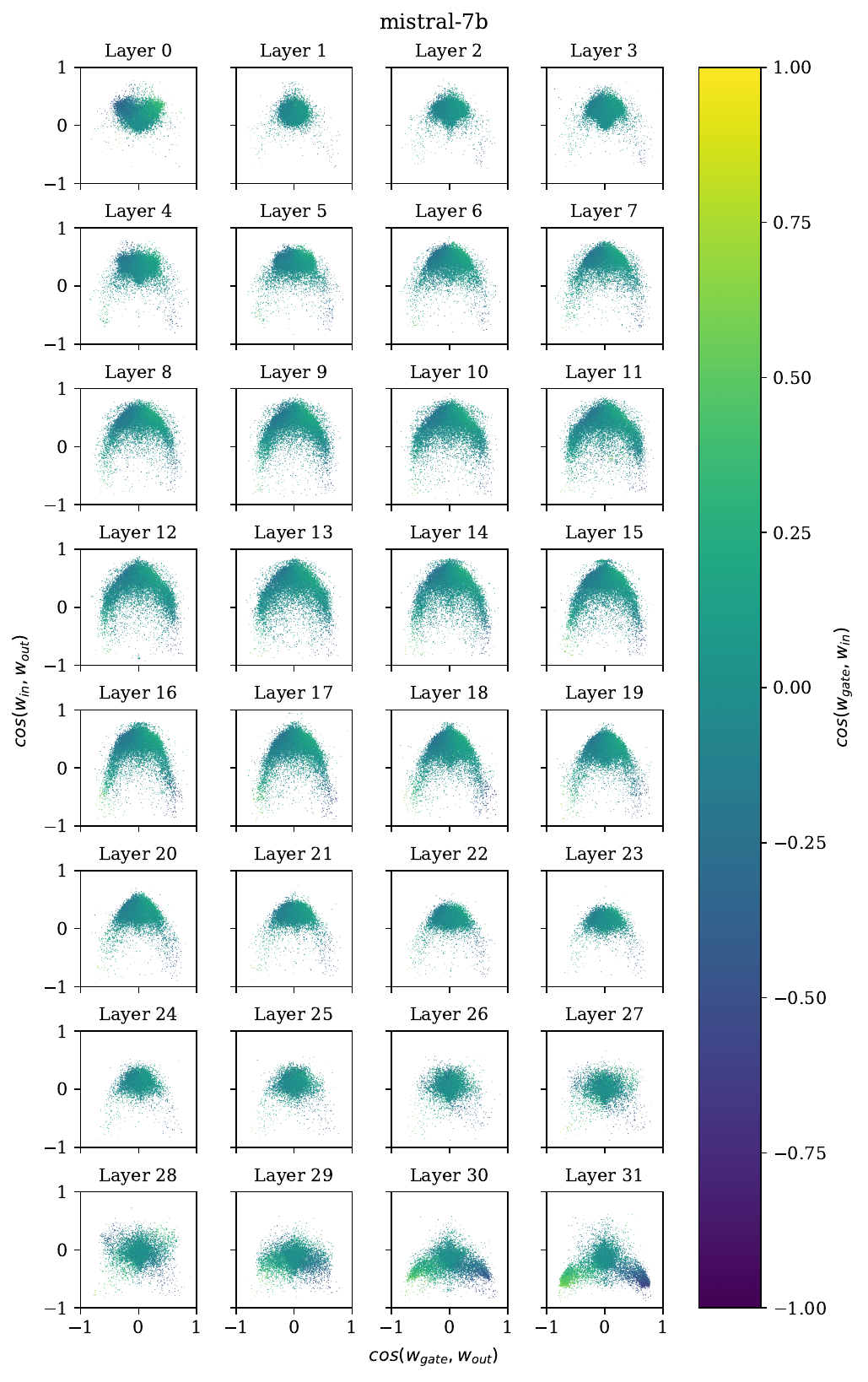}
	\vskip -0.2in
	\caption{}
\end{figure*}
\begin{figure*}
	\centering
	\includegraphics[width=.67\textwidth]{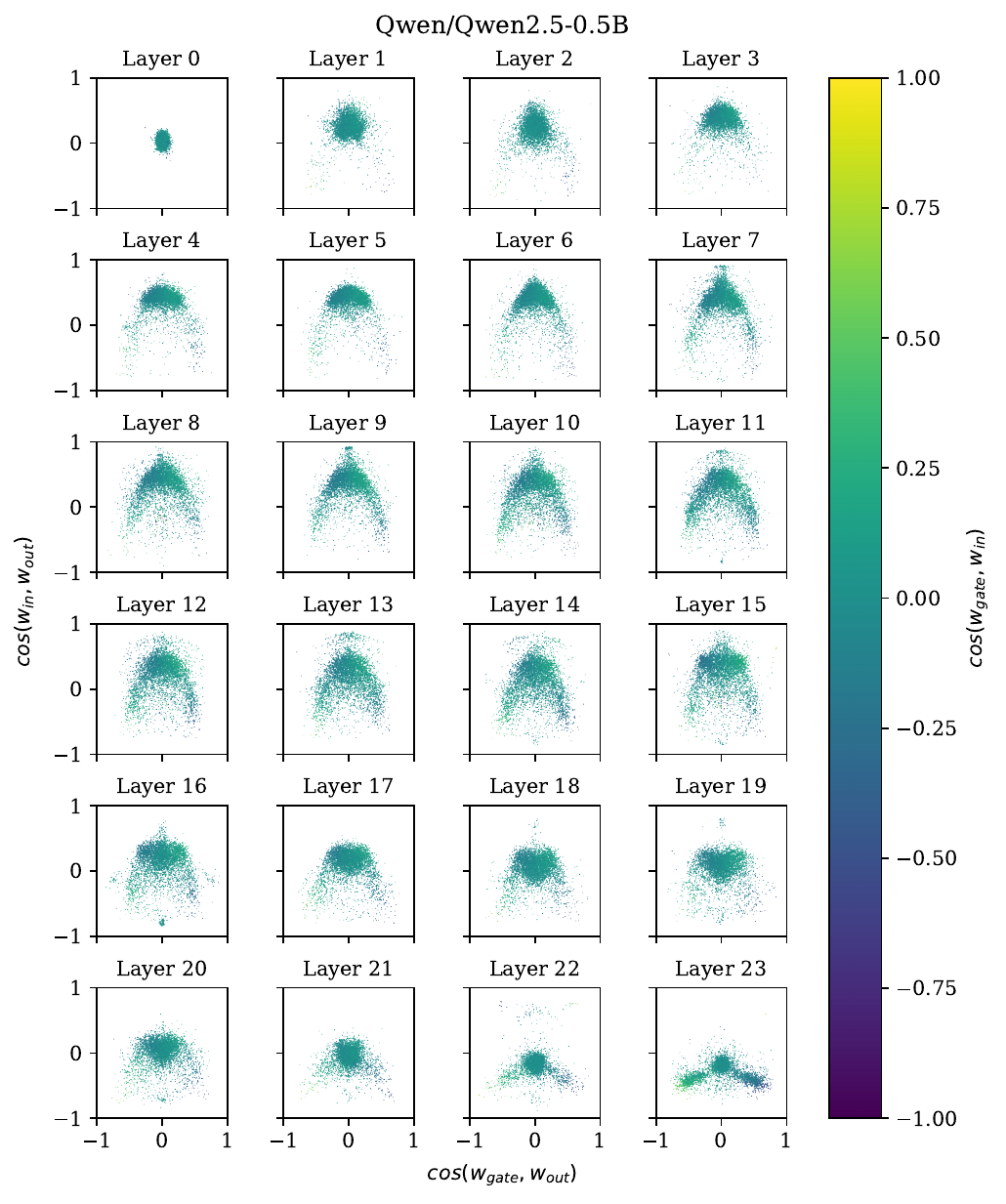}
	\vskip -0.2in
	\caption{}
\end{figure*}
\begin{figure*}
	\centering
	\includegraphics[width=.67\textwidth]{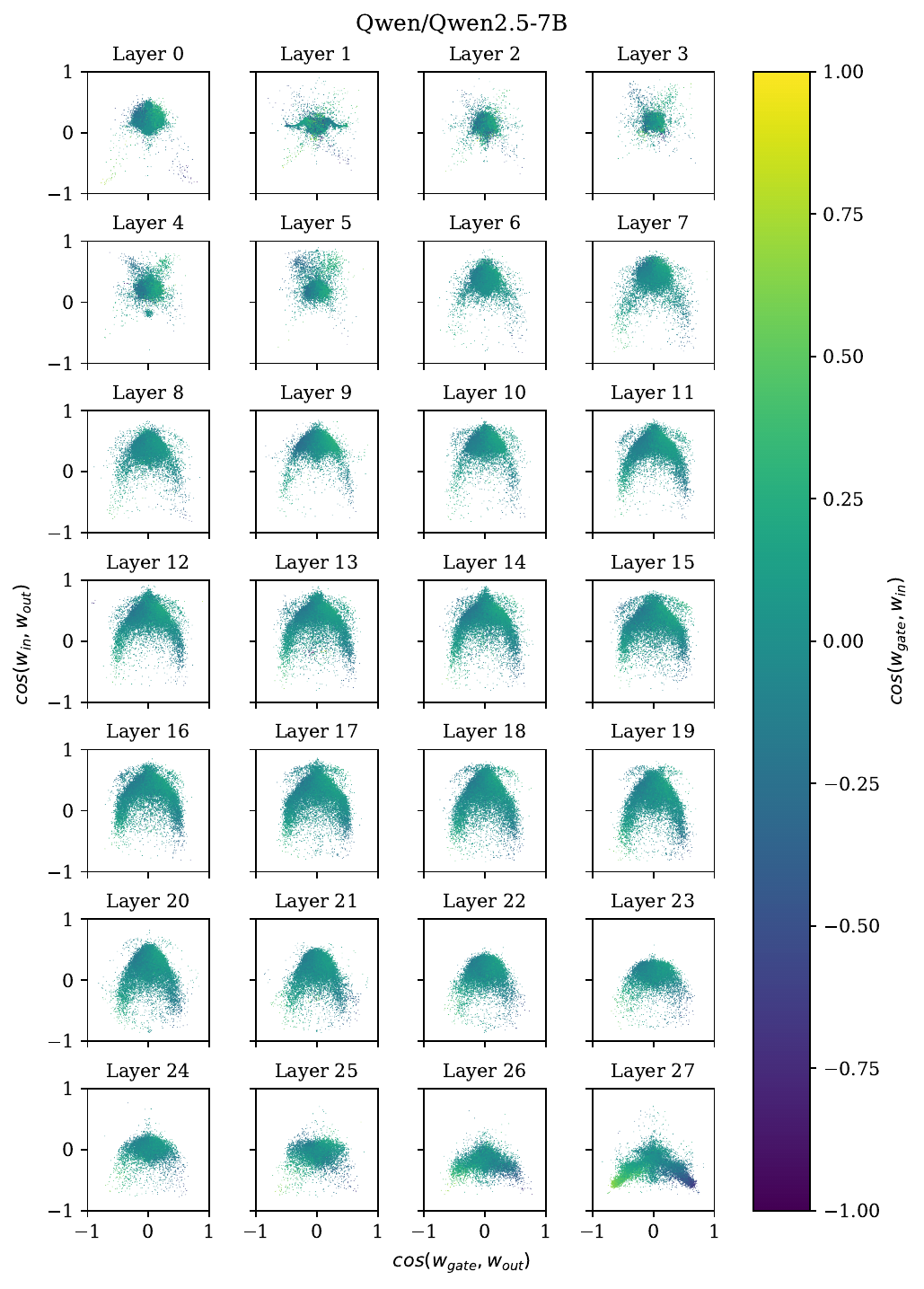}
	\vskip -0.2in
	\caption{}
\end{figure*}
\begin{figure*}
	\centering
	\includegraphics[width=.67\textwidth]{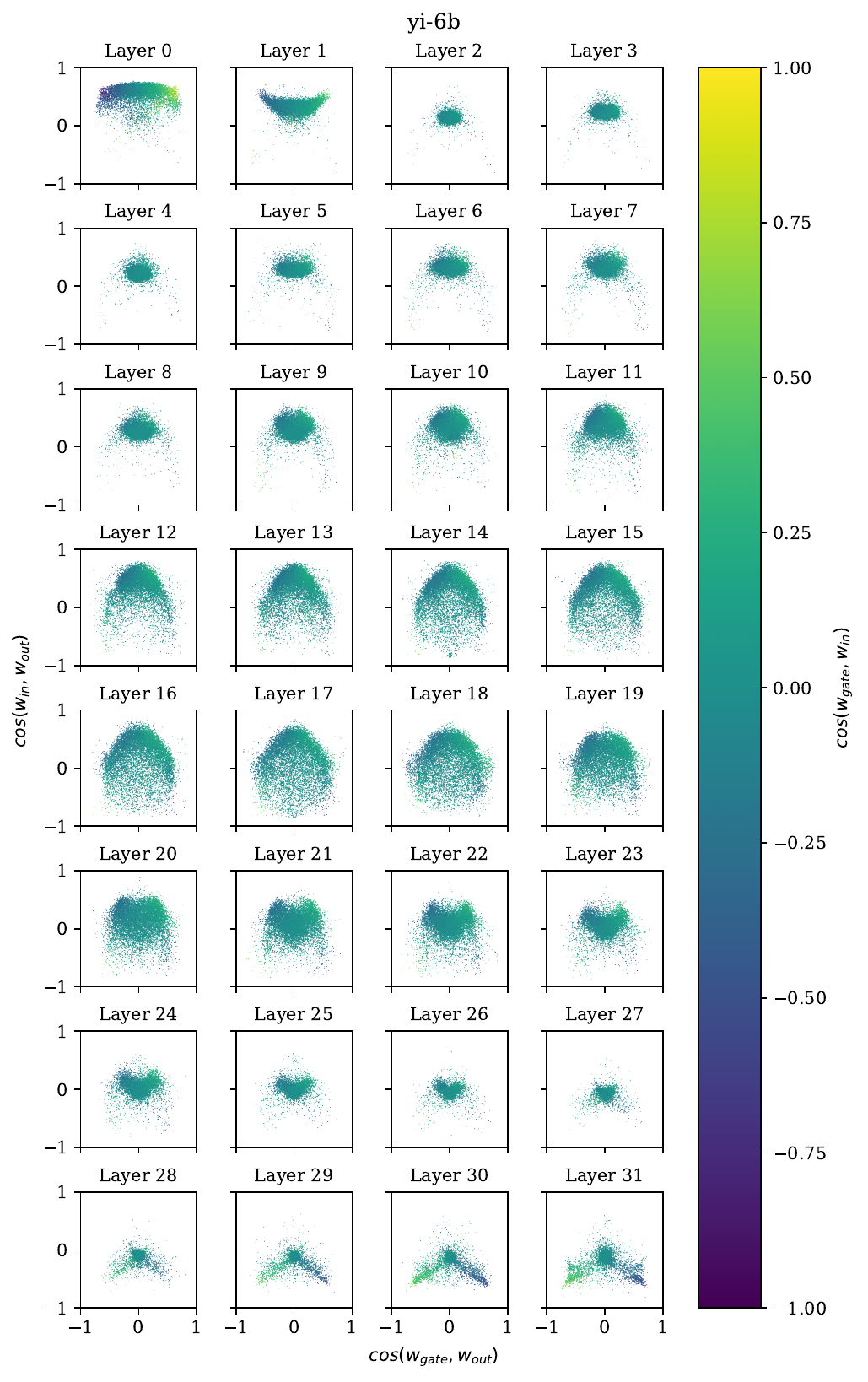}
	\vskip -0.2in
	\caption{}
\end{figure*}

\section{Results across models}\label{ap:hr}
These final figures show our analyses of IO functionalities by layer (\cref{sec:exp}) for all the models we investigated.

We note a few additional patterns that appear only in some of these models:
\begin{itemize}
\item In Yi and the OLMo models, the prevalence of conditional enrichment neurons starts even earlier, at the very first layer.
A particularly interesting example is Yi:
In layer 0 an enormous 68\% of all neurons are conditional enrichment,
then almost none,
then there is a second wave around layers 11-17 (out of 32) which have around 25\% of conditional enrichment neurons each.
\item In some models, especially the OLMo ones, there is a non-negligible number of conditional depletion neurons. They tend to appear in middle-to-late layers, shortly after the conditional enrichment wave.
The clearest example is OLMo-1B, with a peak of 1418 conditional depletion neurons out of 8192 (17\%) in layer 9 out of 16.
\end{itemize}

\end{document}